\documentclass[sigconf]{acmart}

\AtBeginDocument{%
  }

\setcopyright{acmlicensed}
\copyrightyear{2018}
\acmYear{2018}
\acmDOI{XXXXXXX.XXXXXXX}

\acmConference[Conference acronym 'XX]{Make sure to enter the correct
  conference title from your rights confirmation emai}{June 03--05,
  2018}{Woodstock, NY}
\acmISBN{978-1-4503-XXXX-X/18/06}

\usepackage{algorithm}
\usepackage{algpseudocode}
\usepackage{hyperref}
\usepackage{multirow}
\usepackage{makecell}

\copyrightyear{2024}
\acmYear{2024}
\setcopyright{acmlicensed}\acmConference[MM '24]{Proceedings of the 32nd ACM International Conference on Multimedia}{October 28-November 1, 2024}{Melbourne, VIC, Australia}
\acmBooktitle{Proceedings of the 32nd ACM International Conference on Multimedia (MM '24), October 28-November 1, 2024, Melbourne, VIC, Australia}
\acmDOI{10.1145/3664647.3680919}
\acmISBN{979-8-4007-0686-8/24/10}

\begin{document}

\title{Towards High-resolution 3D Anomaly Detection via Group-Level Feature Contrastive Learning
 }
\author{Hongze Zhu}
\orcid{0009-0008-5980-4501}
 \affiliation{%
  \institution{National Engineering Laboratory for Big Data System Computing Technology, Shenzhen University}
  \country{Shenzhen, China}}
 \email{zuhoze@foxmail.com}

 \author{Guoyang Xie}
 \affiliation{%
  \institution{Department of Computer Science, City University of Hong Kong}
  \country{Hong Kong SAR, China}}
      \affiliation{%
  \institution{Department of Intelligent Manufacturing, CATL}
  \country{Ningde, China}
  }
 \email{guoyang.xie@ieee.org}

 \author{Chengbin Hou}
 \affiliation{%
  \institution{Fuzhou Fuyao Institute for Advanced Study, Fuyao University of Science and Technology}
  \country{Fuzhou, China}}
 \email{chengbin.hou10@foxmail.com}

 \author{Tao Dai}
 \affiliation{%
  \institution{College of Computer Science and Software Engineering, Shenzhen University}
  \country{Shenzhen, China}}
 \email{daitao@szu.edu.cn}

\author{Can Gao}
 \affiliation{%
  \institution{College of Computer Science and Software Engineering, Shenzhen University}
  \country{Shenzhen, China}}
 \email{2005gaocan@163.com}

\author{Jinbao Wang}
\orcid{0000-0001-5916-8965}
\authornote{Corresponding author}
 \affiliation{%
  \institution{National Engineering Laboratory for Big Data System Computing Technology, Shenzhen University}
  \country{Shenzhen, China}}
\affiliation{%
  \institution{Guangdong Provincial Key Laboratory of Intelligent Information Processing}
  \country{Shenzhen, China}}
 \email{wangjb@szu.edu.cn}

\author{Linlin Shen}
 \affiliation{%
  \institution{Shenzhen University}
  \country{Shenzhen, China}}
 \affiliation{%
  \institution{Shenzhen Institute of Artificial Intelligence and Robotics for Society}
  \country{Shenzhen, China}}
   \affiliation{%
  \institution{Guangdong Provincial Key Laboratory of Intelligent Information Processing}
  \country{Shenzhen, China}}
 \email{llshen@szu.edu.cn}
\renewcommand{\shortauthors}{Hongze Zhu et al.}

\begin{abstract}
High-resolution point clouds~(HRPCD) anomaly detection~(AD) plays a critical role in precision machining and high-end equipment manufacturing. Despite considerable 3D-AD methods that have been proposed recently, they still cannot meet the requirements of the HRPCD-AD task. There are several challenges: i) It is difficult to directly capture HRPCD information due to large amounts of points at the sample level; ii) The advanced transformer-based methods usually obtain anisotropic features, leading to degradation of the representation; iii) The proportion of abnormal areas is very small, which makes it difficult to characterize. To address these challenges, we propose a novel group-level feature-based network, called Group3AD, which has a significantly efficient representation ability. First, we design an Intercluster Uniformity Network~(IUN) to present the mapping of different groups in the feature space as several clusters, and obtain a more uniform distribution between clusters representing different parts of the point clouds in the feature space. Then, an Intracluster Alignment Network~(IAN) is designed to encourage groups within the cluster to be distributed tightly in the feature space. In addition, we propose an Adaptive Group-Center Selection~(AGCS) based on geometric information to improve the pixel density of potential anomalous regions during inference. The experimental results verify the effectiveness of our proposed Group3AD, which surpasses Reg3D-AD by the margin of 5\% in terms of object-level AUROC on Real3D-AD. We provide
the code and supplementary information on our website:~\href{https://github.com/M-3LAB/Group3AD}{https://github.com/M-3LAB/Group3AD}.
\end{abstract}

\ccsdesc[500]{Computing methodologies~Visual inspection}
\ccsdesc[500]{Computing methodologies~Anomaly detection}

\keywords{Anomaly Detection, 3D Point Clouds, Contrastive Learning, Feature Representation}



\maketitle

\section{Introduction}
Anomaly Detection (AD) is a critical field in machine learning aimed at identifying unusual patterns or abnormalities that do not conform to expected behavior. Traditionally, AD has been extensively applied in 2D image analysis, where methods primarily focus on identifying anomalies through pixel-level discrepancies~\cite{xie2024iad, liu2024deep}. However, these 2D techniques come with inherent limitations, such as the fixed perspectives and inability to capture complex geometries, which often result in loss of important spatial information and context~\cite{li2022towards,liu2024deep,xie2023pushing,xie2024iad,cao2023complementary,bergmann2020uninformed,defard2021padim,Zhu_2024}. Transitioning from 2D images to 3D Point Clouds~(PCD) effectively overcomes these limitations. 3D PCDs provide a more comprehensive representation of objects~\cite{bergmann2021mvtec, guo2020deep}.

Despite the advantages, the field of 3D-AD faces many challenges~\cite{liu2024real3d, chen2023easynet}. The High-Resolution~(HR) of 3D PCDs introduces computational and analytical complexities due to the sheer increase in data volume and the intricacies involved in 3D space analysis~\cite{liu2024real3d, Li_2024_CVPR}. Research based on HRPCD-AD has just emerged, and how to construct efficient representations for AD tasks in HRPCD has become a major challenge. There is an urgent need to improve the precision of HRPCD-AD to meet the needs of industrial manufacturing~(IM)~\cite{liu2024real3d,Li_2024_CVPR}. The obstacle on the road to establishing efficient representation is threefold. i) Existing HRPCD networks necessitate downsampling for large datasets, risking the loss of crucial anomaly information. ii) Recent 3D representation methods yield embeddings with insufficient distinction in AD. iii) The proportion of anomalies in HRPCD is small and obscure, making it easy to overlook~\cite{xie2024iad, liu2024deep, jiang2022softpatch, liu2024real3d, Li_2024_CVPR}.

\begin{figure}[ht]
    \centering
    \includegraphics[width=0.75\linewidth]{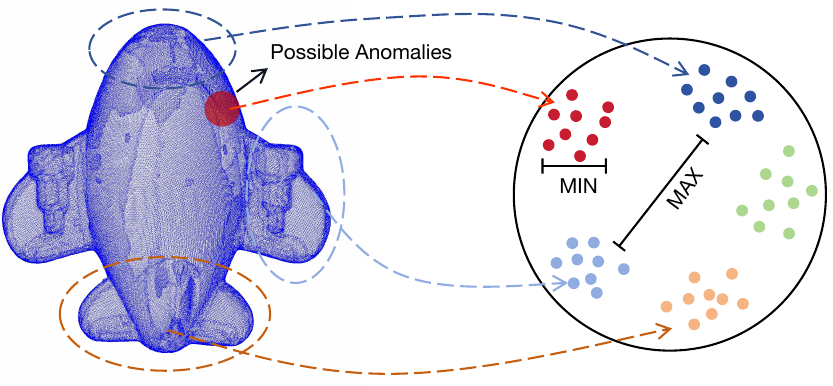}
    \caption{Ideal feature distribution of normal point clouds and abnormalies for the high-resolution 3D-AD task. Group-level features are used to express structural information. 
    }
    \label{fig:ideal_distribution}
\end{figure}


To make the training of the network free from the constraints of HR PCDs and too few training samples, we introduce the idea of group-level feature enhancement. Existing HRPCD networks clearly require downsampling operations when faced with items composed of millions of points during training or inference~\cite{liu2024real3d,Li_2024_CVPR}. Among them, although Reg3D-AD~\cite{liu2024real3d} has a higher group density in HRPCD reasoning in a way similar to PatchCore~\cite{roth2022towards}, it still cannot be directly trained by HRPCD. There is a gap in the middle, which affects the performance. IMRNet also realized the drawbacks of downsampling and proposed adaptive sampling, but still did not fundamentally solve the problem.  We need to study a solution that is not constrained by point cloud resolution. Unlike previous 3D contrastive learning studies~\cite{afham2022crosspoint, liu2023fac,li2022simipu}, we build contrastive learning at group level in a single sample to makes our network training unconstrained by point cloud resolution and scale. The divide and conquer design makes our network's group batchsize infinitely large in theory, although its batchsize is 1.

To achieve better representation, we pose a key question: \textbf{how to create an ideal distribution required by HRPCD-AD in the feature space?} Currently, most network backbones that perform well on HRPCD use the Point Transformer architecture~\cite{liu2024real3d, Li_2024_CVPR}. The embeddings represented by this encoder exhibit anisotropic distribution in the feature space. Research~\cite{gao2019representation, wang2019improving, ethayarajh2019contextual} has shown that transformer-based structures place more emphasis on local contextual information in representation, while ignoring global semantic information. This results in high-frequency groups being distributed in a narrow area and close to the center in the feature space. While low-frequency groups are relatively sparse and far from the origin. Commonly, calculating feature similarity is a widely-used method, but this anisotropic spatial distribution will bring problems to the measurement of feature similarity. Figure~\ref{fig:ideal_distribution} shows an ideal feature space distribution. Specifically, excellent spatial distribution of features should have both uniformity and alignment~\cite{gao2021simcse, cai2020isotropy}. Uniformity requires that the vectors should be distributed as widely as possible and isotropic in the space, while alignment means that the distance between similar vectors in the feature space should be small. Meanwhile, the excellent encoder should be capable of higher anomaly sensitivity to easily distinguish abnormal representation from normal representation in the feature space. Considering above-mentioned distribution characteristics: \textbf{intercluster dispersion} and \textbf{intracluster compactness}, we propose to construct the group-level contrastive learning architecture. The proposed Intercluster Uniformity Network~(IUN) pushes clusters representing similar features far away in the feature space to enhance the uniformity of feature distribution. On the basis of IUN, the Intracluster Alignment Network~(IAN) further tightens the features in the same cluster to strengthen the alignment of feature distribution.

To better capture subtle features, we design the Adaptive Group-Center Selection~(AGCS). Drawing inspiration from the practices of quality inspection engineers, who often focus more intensively on areas they suspect to be problematic to enhance the detection of abnormalities, AGCS similarly prioritizes regions within 3D point clouds that exhibit potential anomalies. Leveraging the Fast Point Feature Histogram (FPFH)~\cite{rusu2009fast}, AGCS intensifies the examination of areas with significant local geometric variations. Finally, AGCS enables our network to preserve more potential anomaly groups during inference, thereby enhancing its AD capability.


Our contributions are succinctly outlined as follows:
\begin{itemize}
    \item A novel Group3AD framework for the HRPCD-AD task is designed, which optimizes the spatial information of 3D point clouds, enhancing the precision of anomaly detection.
    \item We propose the Intercluster Uniformity Network~(IUN) and the Intracluster Alignment Network~(IAN), which can disperse features across clusters and tighten features within clusters in the feature space, respectively. Both two networks enhance feature uniformity and alignment, improving the coherence of feature representations for anomaly identification.
    \item An efficient Adaptive Group-Center Selection~(AGCS) is designed, which focuses on regions with potential anomalies, enhancing model sensitivity and detection efficiency.
    \item The proposed Group3AD is flexible and scalable. Group3AD can be directly integrated with other network architectures, which promotes the wider application of various anomaly detection tasks.
\end{itemize}

\section{RELATED WORK}


\subsection{2D Anomaly Detection}
Since the introduction of the MVTec AD dataset~\cite{bergmann2019mvtec}, 2D image anomaly detection (2D-AD) has garnered increased focus~\cite{xie2024iad,liu2024deep}. Predominantly, research in this domain utilizes this 2D dataset for exploring unsupervised AD techniques \cite{li2021cutpaste,liu2023simplenet,bergmann2020uninformed,deng2022anomaly,gudovskiy2022cflow,rudolph2021same,defard2021padim,roth2022towards,zavrtanik2021draem,wang2021student,schluter2022natural,zavrtanik2022dsr,gudovskiy2022cflow,rudolph2021same}. 
The intention of image reconstruction-based 2D-AD methods is to reconstruct abnormal images into approximate normal images and achieve anomaly localization through pixel-level comparison. The feature extraction based 2D-AD methods strives to provide more informative embeddings, with more significant differences between normal and anomalous features. Due to the fact that many networks in the former use the method of training from scratch, their performance may be inferior to that of the latter, which uses robust pre-training models.

Notable examples include DRAEM~\cite{zavrtanik2021draem}, SimpleNet~\cite{liu2023simplenet}, etc. They create fake samples and identify anomalies through reconstruction comparison, building supervised tasks in unsupervised datasets. The representation of pre-trained networks has been proven to be more powerful and effective. PatchCore~\cite{roth2022towards} employs a memory-efficient representation of normal data distribution through sparse sampling of feature space patches for identifying outliers. CutPaste~\cite{li2021cutpaste} markedly enhances AD capabilities by artificially generating anomalies through cutting and pasting segments within images. Self-Taught AD~\cite{bergmann2020uninformed} leverages a student-teacher framework where features from a pre-trained teacher network guide a student network trained on normal data to detect anomalies based on feature discrepancies. STPM~\cite{wang2021student} and MKD~\cite{salehi2021multiresolution} both employ a teacher-student architecture for AD, with STPM harnessing multi-scale features directly, while MKD concentrates on distilling knowledge from these multi-scale features through a more efficient network architecture to enhance performance. Normalizing flow~\cite{lei2023pyramidflow} methods have created a feature distribution with anomaly sample deviation, expanding its potential.
Although 2D-AD methods cannot be directly used for 3D-AD, they highlight the importance of building an easily distinguishable feature distribution and enhancing the representational ability of the network~\cite{xie2024iad,liu2024deep}.

\subsection{RGB-D Anomaly Detection}
Despite significant advancements in 2D image anomaly detection, the exploration of anomaly detection in 3D PCDs remains relatively underdeveloped~\cite{bergmann2023anomaly,bonfiglioli2022eyecandies,chen2023easynet}. This field saw a notable surge in interest following the release of the pioneering real-world 3D anomaly detection dataset, known as MVTec 3D-AD~\cite{bergmann2021mvtec}. The key challenge in 3D-AD lies in the effective harnessing of depth information, which can significantly enhance detection capabilities in certain scenarios.

While AST~\cite{rudolph2023asymmetric} achieves good results on the MVTec 3D-AD by utilizing depth information to separate the background, it primarily depends on 2D-AD techniques for anomaly detection, neglecting the depth characteristics of objects. Designing the feature extraction module for detecting anomalies in PCD demands fresh approaches. Recent studies have made efforts to design networks with stronger point clouds representation capabilities~\cite{cao2023complementary}. 3D-ST~\cite{bergmann2023anomaly} introduces a self-supervised learning strategy for representation learning in PCD and employs a knowledge distillation-based modeling module. While CPMF~\cite{cao2023complementary} streamlines anomaly detection by projecting point clouds into 2D images from various angles to reduce feature extraction complexity and computational costs, this approach does not fully leverage the advantages inherent in the 3D nature of point clouds data. BTF~\cite{horwitz2023back} underscores the efficacy of traditional handcrafted PCD descriptors, yet notes the underperformance of learned features. BTF attributes this paradox to the inadequate transferability of existing pretrained features on the current small-scale PCD datasets. The use of spatial information for AD tasks still deserves further exploration~\cite{Li_2024_CVPR,wang2023multimodal,cao2023complementary,liu2024real3d}.

\subsection{High-resolution 3D Anomaly Detection}
Some methods have made efforts in the direction of PCD anomaly representation. However, MVTec 3D-AD~\cite{bergmann2021mvtec} is an RGB-D dataset with low resolution, which cannot further explore the value of spatial information in AD tasks. The proposal of the Real3D-AD~\cite{liu2024real3d} dataset containing HR multi-view information of objects brings more development space to AD. The point resolution and precision of Real3D-AD are 4.28 and 9 times higher than MVTec 3D-AD, respectively. The ultra-high accuracy brings more potential and hope to AD tasks but poses significant challenges to establishing representations. CPMF~\cite{cao2023complementary} aims to achieve PCD AD by merging handcrafted PCD descriptions with the capabilities of pre-trained 2D neural networks. CPMF performs well on the RGBD dataset, but poorly on Real3D-AD. Reg3D-AD~\cite{liu2024real3d} is a benchmark method built on Real3D-AD, which is built on the memory bank and achieves significant performance improvements using pre-trained models. However, the pre-training of Reg3D-AD is difficult to establish directly on HRPCD datasets. IMRNet~\cite{Li_2024_CVPR} introduces a self-supervised, scalable framework for 3D PCD AD, leveraging iterative mask reconstruction and geometry-aware sampling to identify and localize anomalies with high accuracy. However, IMRNet has also been affected by the negative performance impact of downsampling.

In view of this, our work aims to design a method that fully utilizes the HRPCD spatial information to enhance the model's generalization ability with small samples in unsupervised manner.

\begin{figure*}[th]
    \centering
    \includegraphics[width=1\linewidth]{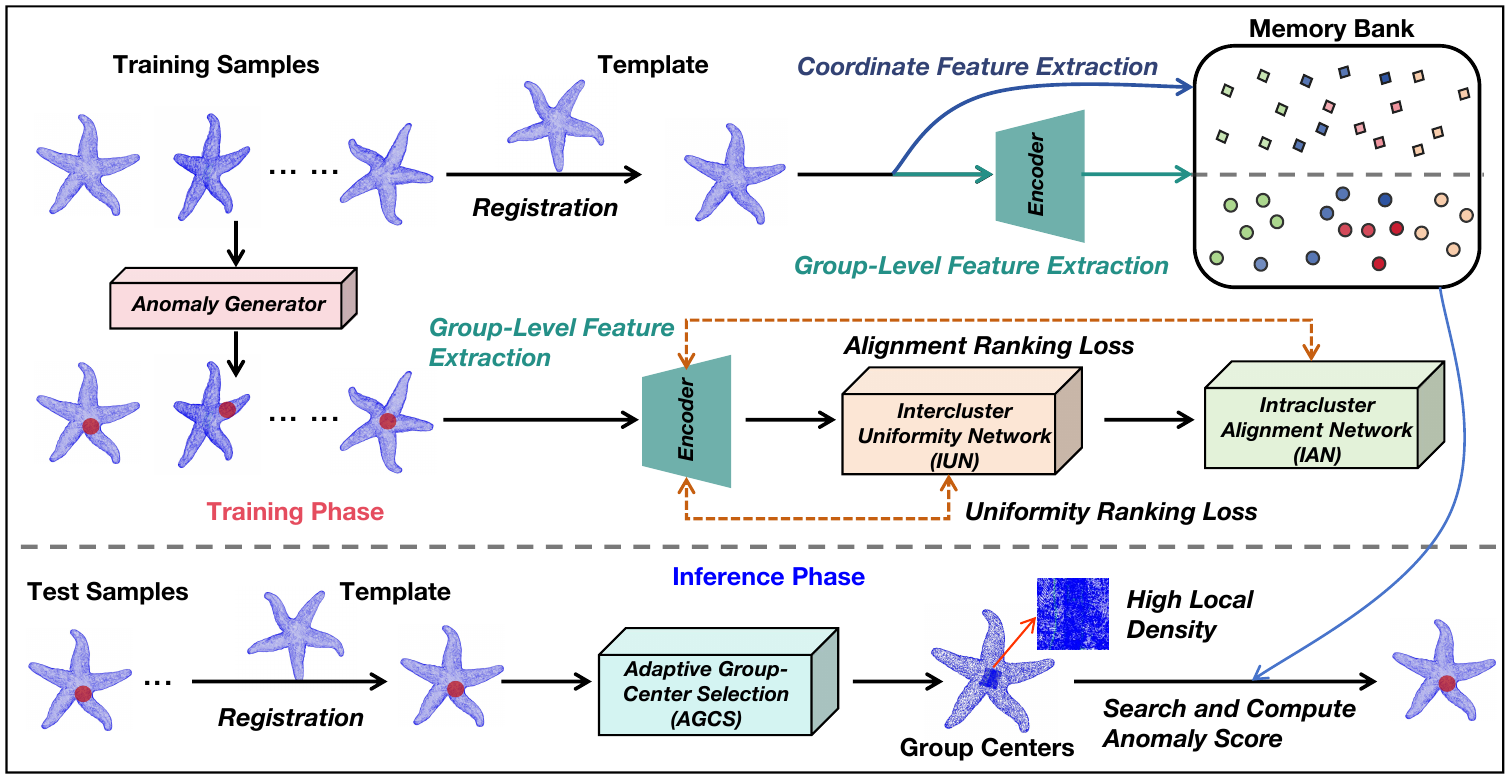}
    \caption{The pipeline of Group3AD, which consists of three main components. (1) Group-Level Feature Extraction extracts group-level features from the input 3D point clouds. (2) Intercluster Uniformity Network (IUN) and Intracluster Alignment Network (IAN) enhance the feature separation between clusters and tighten the distribution within clusters, respectively, for improving anomaly detection accuracy. (3) Adaptive Group-Center Selection (AGCS), used during inference, dynamically focuses on regions with potential anomalies by adjusting the sampling density based on geometric information. This structured approach ensures efficient anomaly detection in complex 3D environments.}
    \label{fig:group3ad_pipeline}
\end{figure*}

\section{APPROACH}
\subsection{Problem Definition}

Our approach to 3D-AD aligns with the settings defined by Real3D-AD~\cite{liu2024real3d}. Defined formally, our task involves a set of training examples $T = \{t_i\}_{i=1}^N$, where each $t_i$ is a normal point clouds sample, belonging to a specific category $c_j$, with $c_j \in C$ where $C$ represents the entire set of categories. In testing, determine whether a sample in a certain category contains anomalies, and if so, define the entire sample as an anomaly and locate the anomaly area.


\subsection{Group3AD}
 This section delves into the detailed architecture of Group3AD, as shown in Figure~\ref{fig:group3ad_pipeline}, a model specifically designed to enhance the resolution and accuracy of 3D anomaly detection through group-level feature contrastive learning.

\subsubsection{Intercluster Uniformity Network~(IUN)}
The IUN establishes an intercluster contrastive learning task. Each cluster is composed of group-level feature vectors with similarity. This network sets each cluster as negative samples in contrastive learning, and widens the distance between each cluster to obtain a uniform structured sample space.

We construct fake anomalies on the original HR point clouds so that the encoder can still maintain uniformity when encoding anomaly groups during inference. Given the original point clouds $P$, $P_i$ represents the $i$th point in $P$. Firstly, calculate the FPFH features of the point clouds:
\begin{equation}
FPFH(P_i) = ComputeFPFH(P_i,\text{neighbors}(P_i)).
\end{equation}
Next, select the point$c$ with the highest FPFH feature as the center of the local region:
\begin{equation}
c = \text{argmax}_{p_i}FPFH(P_i).
\end{equation}
Then, based on the center point$c$, a random $r\in[1\%,10\%]$ of the total point clouds will be selected around c as the local area:
\begin{equation}
P_{local} = \{ P_j\mid \text{distance}(P_j,c)\leq d,j = 1,...,N \},
\end{equation}
where $d$ is a distance threshold that ensures $P_{local}$ contains approximately $r$ of the points in $P$ are included. Subsequently, randomly select a value with a variance $\sigma^2 \in [0.01,0.05]$ and use $\sigma^2$ to generate a normally distributed random noise $\epsilon$ with an average value of 0. Add noise to each dimension of $P$ to generate anomaly regions $P_{anomaly}$:
\begin{equation}
P_{\text{anomaly},i} = P_{\text{local},i} + \epsilon, 
\end{equation}
where $\epsilon = \mathcal N(0,\sigma^2)$.
Repeat this process for all $P$ midpoints to obtain the set of anomalies $A$:
\begin{equation}
A ={  P_{\text{anomaly},i}\mid\forall P_{\text{local},i}\in P_{\text{local}} }.
\end{equation}
Combine $A$ and $P$ to obtain $FakeAnomalyP$:
\begin{equation}
FakeAnomaly(P) = P\cup A.
\end{equation}

We cluster group-level features using the K-means~\cite{macqueen1967some} algorithm to determine which cluster ($P_i$) belongs to. K-means determines the number of clusters using the Elbow Method~\cite{syakur2018integration}. Determine the final size of $K$ by plotting the relationship between Within Cluster Sum of Squares~(WCSS) and the number of clusters ($K$). Find the "Elbow" point in the relationship diagram, which is the position where the rate of WCSS decrease suddenly slows down as $k$ increases. This point is usually considered the optimal number of clusters, as it represents the best balance between the intracluster compactness and the number of clusters. 

The calculation of WCSS is represented as:
\begin{equation}
W_k = \sum_{x_i\in C_k}\left\|X_i- \mu_k \right\|_2^{2},
\end{equation}
where $\mu_k$ is the $k$th cluster center, $C_k$ is the point set in the $k$th cluster, $W_k$ is the WCSS of the $k$th cluster, and $\parallel X_i - \mu_k \parallel^2$ is the square of the Euclidean distance from point $x_i$ to the cluster center $\mu_k$. The entire dataset $W$ is the sum of all clustered WCSS:
\begin{equation}
W=\sum_{k=1}^{K} W_{k}.
\end{equation}

\begin{figure}[ht]
    \centering
    \includegraphics[width=0.85\linewidth]{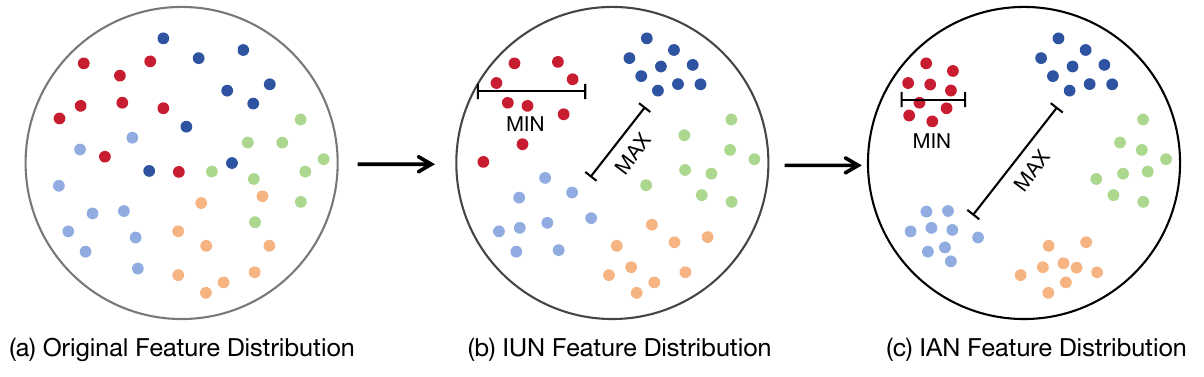}
    \caption{Flowchat of group-level feature distribution, constrained by IUN and IAN. The basic idea is to minimize the intra-group distance and maximize the inter-group distance.}
    \label{fig:feature_distribution}
\end{figure}

Assuming we have $k$ cluster centers, denoted as $C =\{C_1,C_2,...,C_k\}$, where each cluster center $C_i$ is an $n$-dimensional vector. For any two different cluster centers $C_i$ and $C_j$, we calculate the Euclidean distance between them:
\begin{equation}
d\left(C_{i}, C_{j}\right)=\left\|C_{i}-C_{j}\right\|_2,
\end{equation}
where $\|\cdot\|$ represents calculating Euclidean distance. We compare the distances between each pair of cluster centers and select the minimum value among these distances:
\begin{equation}
\min \_ \text {dist}=\min _{1 \leq i<j \leq K} \text{distance}\left(C_{i}, C_{j}\right).
\end{equation}
Define the uniformity ranking loss function as the reciprocal of this minimum distance:
\begin{equation}
L_{uniformity} =\frac{1}{{ \min \_ \text {dist} }}.
\end{equation}
The larger the value of the uniformity ranking loss function, the smaller the minimum distance between cluster centers, the lower the discrimination between clusters, and the greater the loss. Therefore, our goal is to minimize this loss function, thereby maximizing the minimum distance between cluster centers and improving the quality of clustering. The changes in the distribution of vectors in the feature space represented by the optimized encoder through the uniformity ranking loss are shown in Figure~\ref{fig:feature_distribution}(b).

\subsubsection{Intracluster Alignment Network~(IAN)} The IAN establishes an intracluster contrastive learning task. Each cluster constitutes one mini-batch. Each vector in the cluster constitutes one positive sample in contrastive learning. Assuming we have $k$ cluster, denoted as $C = \{C_1,C_2,...,C_k\}$, where each cluster $C_k$ contains a certain number of points. We use $P_{k}=\left\{P_{k 1}, P_{k 2}, \ldots, P_{k n}\right\}$ to represent the set of points $C_k$ contains. For any two points $P_{ki}$ and $P_{kj}$ in cluster $C_k$, we calculate the Euclidean distance between them:
\begin{equation}
d\left(P_{k i}, P_{k j}\right)=\left\|P_{k i}-P_{k j}\right\|,
\end{equation}
where $\|\cdot\|$ represents calculating Euclidean distance. For each cluster $C_k$, we need to find the maximum distance between all pairs of internal points:
\begin{equation}
\max \_d i s t_{k}=\max _{P_{k i}, P_{k j} \in P_{k}} d\left(P_{k i}, P_{k j}\right),
\end{equation}
This means that we compare the distances between all possible point pairs within cluster $C_k$ and select the maximum value among these distances. Finally, we define the alignment ranking loss function as the average of the maximum intracluster distance of all clusters:
\begin{equation}
L_{\text {alignment }}=\frac{T}{k} \sum_{k=1}^{K} \max {\_} \text {dist}_{k}^{2},
\end{equation}
where $T$ is a temperature coefficient that makes the training of the network more stable. A larger intracluster distance usually means that the points within the cluster are more dispersed. The larger the value of this loss function, the more scattered the points within the cluster, the lower the compactness of the cluster, and the greater the loss. As shown in the Figure~\ref{fig:feature_distribution}(c), the goal of IAN is to minimize this loss function and improve the compactness of clustering.
\subsubsection{Adaptive Group-Center Selection~(AGCS)}

\begin{figure}[ht]
    \centering
    \includegraphics[width=0.6\linewidth]{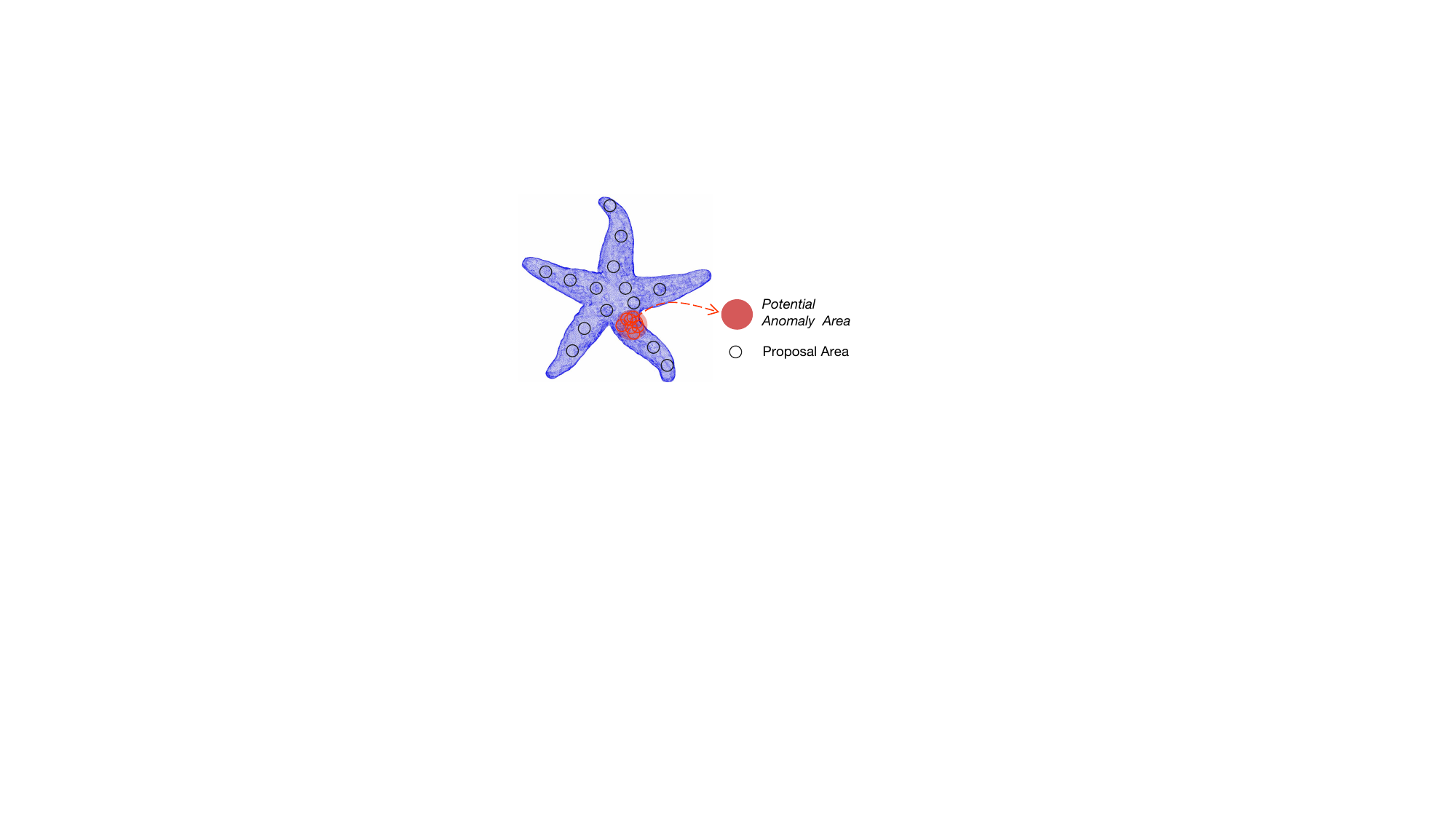}
    \caption{Adaptive group-center selection~(AGCS). AGCS adaptively selects the points most likely to be within the anomaly region in the selection of group centers via FPFH features.}
    \label{fig:group_center}
\end{figure}

The detailed algorithm and parameter settings of AGCS for ablation experiments can be found in the supplementary materials. Although our method does not require sampling, we still need to use the Farthest Point Sampling~(FPS) algorithm to select the center points of the groups. As shown in the Figure~\ref{fig:group_center}, the goal of AGCS is to allocate higher groups resolution in areas with significant geometric changes in inference phase. 

To evaluate the geometric characteristics of points, we employ the Fast Point Feature Histogram~(FPFH) approach.
For each point $ p$ within the PCD $P$, We calculate the surface normal of this point, denoted as $n(\cdot)$. We take $n$ neighboring points around point $p$, denoted as ${\{N_1,N_2,...,N_n\}}$. We calculate the surface normal difference between point $p$ and its $n$ neighboring points through Euclidean distance, and obtain the surrounding variation feature $F$. We have:
\begin{equation}
F_p = \sum_{i=1}^{N} \left\| f(p) - f(N_i) \right\|_2.
\end{equation}
This metric enables us to quantify the dissimilarity in geometric attributes between a point and its neighboring points based on their FPFH descriptors. In the sampling process of center points, it is assumed that a total of$n$ center points are required.
The center points $C_{FPFH}$ sampled by the FPFH method are $a\times n$, and the remaining $(1-a)\times n$ center points $C_{FPS}$ are obtained through FPS sampling, where $a$ represents the assigned weight. Ultimately, all center points $C$ are composed of both:
\begin{equation}
C = C_{FPFH}\cup C_{FPS}.
\end{equation}

\subsubsection{Overall Pipeline}

In the Group3AD framework, the \textbf{training phase} starts with a pre-trained encoder to extract features from 3D point clouds initially. The Intercluster Uniformity Network~(IUN) subsequently enhances the encoder's capabilities, which increases the separation between different clusters to improve the distinctness of feature representations. Additionally, the Intracluster Alignment Network~(IAN) further refines these features by ensuring they are tightly aligned within each cluster. The enhanced encoder then populates a memory bank with these optimized features.  In addition, a memory bank based on coordinate information, similar to Reg3D-AD~\cite{liu2024real3d}, is used to assist in detection. During the \textbf{inference phase}, the Adaptive Group-Center Selection~(AGCS) assists by selectively targeting potential anomaly regions, comparing these regions against the memory bank to accurately compute anomaly scores of groups. The interpolation method for obtaining anomaly scores for each point in this paper is the same as Reg3D-AD~\cite{liu2024real3d}.

The training and inference implementation of Group3AD are described in Algorithm~\ref{alg:Group3AD}.
We denote $\mathbf{P}$ as a set of point clouds from the training or testing dataloader, and use $\mathbf{P}_{\text{mod}}$ for the modified point clouds after generating fake anomalies, $\mathbf{F}$ for features extracted by the encoder $\Phi_{\text{enc}}$.
$C$ and $\mathbf{C}_{\text{centers}}$ are for clusters and their centers, respectively.
$\mathcal{L}_{\text{IUN}}$ and $\mathcal{L}_{\text{IAN}}$ are for the loss functions of the IUN and IAN phases.
$\mathbf{C}_{\text{points}}$ is for the center points selected by AGCS.
$\mathbf{S}$ is for the computed anomaly scores during inference. 

\begin{algorithm}[th]
\caption{Group3AD with Two-Phase Training}\label{alg:Group3AD}
\begin{algorithmic}[1]
\State \textbf{Input:} Training dataloader $\mathcal{D}_{\text{train}}$, Testing dataloader $\mathcal{D}_{\text{test}}$, epochs $E$
\State \textbf{Output:} Trained Encoder $\Phi_{\text{enc}}$, IUN $\Phi_{\text{IUN}}$, IAN $\Phi_{\text{IAN}}$, and AGCS settings
\State Initialize $\Phi_{\text{enc}}$, $\Phi_{\text{IUN}}$, and $\Phi_{\text{IAN}}$ randomly
\State \textcolor{red}{/*Phase 1: Training IUN*/}
\For{$i = 1$ to $E/2$}
    \For{Pointclouds $\mathbf{P}$ from $\mathcal{D}_{\text{train}}$}
        \State $\mathbf{P}_{\text{mod}} \gets$ generate\_fake\_anomalies($\mathbf{P}$)
        \State $\mathbf{F} \gets \Phi_{\text{enc}}(\mathbf{P}_{\text{mod}})$
        \State $C, \mathbf{C}_{\text{centers}} \gets$ cluster\_features($\mathbf{F}$)
        \State $\mathcal{L}_{\text{IUN}} \gets$ compute\_IUN\_loss($\mathbf{C}_{\text{centers}}$)
        \State Perform backpropagation on $\mathcal{L}_{\text{IUN}}$
    \EndFor
\EndFor
\State \textcolor{red}{/*Phase 2: Training IAN*/}
\For{$i = E/2 + 1$ to $E$}
    \For{Pointclouds $\mathbf{P}$ from $\mathcal{D}_{\text{train}}$}
        \State $\mathbf{F} \gets \Phi_{\text{enc}}(\mathbf{P})$
        \State $C \gets$ cluster\_features($\mathbf{F}$) \Comment{Assuming the function also returns clusters $C$}
        \State $\mathcal{L}_{\text{IAN}} \gets$ compute\_IAN\_loss($C$)
        \State Perform backpropagation on $\mathcal{L}_{\text{IAN}}$
    \EndFor
\EndFor
\State \textcolor{blue}{/*Inference Phase*/}
\For{Pointclouds $\mathbf{P}$ from $\mathcal{D}_{\text{test}}$}
    \State $\mathbf{C}_{\text{points}} \gets$ AGCS.select\_center\_points($\mathbf{P}$)
    \State $\mathbf{F} \gets \Phi_{\text{enc}}(\mathbf{P})$
    \State $\mathbf{S} \gets$ compute\_anomaly\_scores($\mathbf{F}$, $\mathbf{C}_{\text{points}}$)
    \Comment{Further processing based on $\mathbf{S}$}
\EndFor
\end{algorithmic}
\end{algorithm}

\begin{table*}[th]
\caption{O-AUROC score ($\uparrow$) on Real3D-AD. The best and second-best results are marked in red and blue, respectively.}
\resizebox{\textwidth}{!}{
\begin{tabular}{l|cccccccccccc|c}
\toprule
\textbf{Method} &
  \multicolumn{1}{l}{\textbf{Airplane}} &
  \multicolumn{1}{l}{\textbf{Car}} &
  \multicolumn{1}{l}{\textbf{Candybar}} &
  \multicolumn{1}{l}{\textbf{Chicken}} &
  \multicolumn{1}{l}{\textbf{Diamond}} &
  \multicolumn{1}{l}{\textbf{Duck}} &
  \multicolumn{1}{l}{\textbf{Fish}} &
  \multicolumn{1}{l}{\textbf{Gemstone}} &
  \multicolumn{1}{l}{\textbf{Seahorse}} &
  \multicolumn{1}{l}{\textbf{Shell}} &
  \multicolumn{1}{l}{\textbf{Starfish}} &
  \multicolumn{1}{l|}{\textbf{Toffees}} &
  \multicolumn{1}{l}{\textbf{Average}} \\ \hline
BTF(Raw) &
  0.520 &
  0.560 &
  0.462 &
  0.432 &
  0.545 &
  {\color[HTML]{FE0000} 0.784} &
  0.549 &
  0.648 &
  0.779 &
  {\color[HTML]{FE0000} 0.754} &
  {\color[HTML]{3531FF} 0.575} &
  0.630 &
  0.603 \\
BTF(FPFH) &
  0.730 &
  0.647 &
  0.703 &
  0.789 &
  0.707 &
  {\color[HTML]{3531FF} 0.691} &
  0.602 &
  {\color[HTML]{FE0000} 0.686} &
  0.596 &
  0.396 &
  0.530 &
  0.539 &
  0.635 \\
M3DM(PointBERT) &
  0.407 &
  0.506 &
  0.442 &
  0.673 &
  0.627 &
  0.466 &
  0.556 &
  0.617 &
  0.494 &
  0.577 &
  0.528 &
  0.562 &
  0.538 \\
M3DM(PointMAE) &
  0.434 &
  0.541 &
  0.450 &
  0.683 &
  0.602 &
  0.433 &
  0.540 &
  0.644 &
  0.495 &
  {\color[HTML]{3531FF} 0.694} &
  0.551 &
  0.552 &
  0.552 \\
PatchCore(FPFH) &
  {\color[HTML]{FE0000} 0.882} &
  0.590 &
  0.565 &
  0.837 &
  0.574 &
  0.546 &
  0.675 &
  0.370 &
  0.505 &
  0.589 &
  0.441 &
  0.541 &
  0.593 \\
PatchCore(FPFH+Raw) &
  {\color[HTML]{3531FF} 0.848} &
  {\color[HTML]{FE0000} 0.777} &
  0.626 &
  {\color[HTML]{FE0000} 0.853} &
  0.784 &
  0.628 &
  0.837 &
  0.359 &
  0.767 &
  0.663 &
  0.471 &
  0.570 &
  0.682 \\
PatchCore(PointMAE) &
  0.726 &
  0.498 &
  0.585 &
  0.827 &
  0.783 &
  0.489 &
  0.630 &
  0.374 &
  0.539 &
  0.501 &
  0.519 &
  0.663 &
  0.594 \\
CPMF &
  0.632 &
  0.518 &
  0.718 &
  0.640 &
  0.640 &
  0.554 &
  0.840 &
  0.349 &
  {\color[HTML]{FE0000} 0.843} &
  0.393 &
  0.526 &
  {\color[HTML]{FE0000} 0.845} &
  0.625 \\
IMRNet &
  0.762 &
  0.711 &
  0.755 &
  0.780 &
  {\color[HTML]{3531FF} 0.905} &
  0.517 &
  0.880 &
  {\color[HTML]{3531FF} 0.674} &
  0.604 &
  0.665 &
  {\color[HTML]{FE0000} 0.674} &
  0.774 &
  {\color[HTML]{3531FF} 0.725} \\
\textbf{Reg3D-AD} &
  0.716 &
  0.697 &
  {\color[HTML]{3531FF} 0.827} &
  {\color[HTML]{3531FF} 0.852} &
  0.900 &
  0.584 &
  {\color[HTML]{3531FF} 0.915} &
  0.417 &
  0.762 &
  0.583 &
  0.506 &
  0.685 &
  0.704 \\
\textbf{Group3AD(Ours)} &
  0.744 &
  {\color[HTML]{3531FF} 0.728} &
  {\color[HTML]{FE0000} 0.847} &
  0.786 &
  {\color[HTML]{FE0000} 0.932} &
  0.679 &
  {\color[HTML]{FE0000} 0.976} &
  0.539 &
  {\color[HTML]{3531FF} 0.841} &
  0.585 &
  0.562 &
  {\color[HTML]{3531FF} 0.796} &
  {\color[HTML]{FE0000} 0.751} \\  \bottomrule

\end{tabular}}
\label{table1} 
\end{table*}

\begin{table*}[th]
\caption{O-AUPR score ($\uparrow$) on Real3D-AD. The best and second-best results are marked in red and blue, respectively.}
\resizebox{\textwidth}{!}{
\begin{tabular}{l|cccccccccccc|c}
\toprule
\textbf{Method} &
  \multicolumn{1}{l}{\textbf{Airplane}} &
  \multicolumn{1}{l}{\textbf{Car}} &
  \multicolumn{1}{l}{\textbf{Candybar}} &
  \multicolumn{1}{l}{\textbf{Chicken}} &
  \multicolumn{1}{l}{\textbf{Diamond}} &
  \multicolumn{1}{l}{\textbf{Duck}} &
  \multicolumn{1}{l}{\textbf{Fish}} &
  \multicolumn{1}{l}{\textbf{Gemstone}} &
  \multicolumn{1}{l}{\textbf{Seahorse}} &
  \multicolumn{1}{l}{\textbf{Shell}} &
  \multicolumn{1}{l}{\textbf{Starfish}} &
  \multicolumn{1}{l|}{\textbf{Toffees}} &
  \multicolumn{1}{l}{\textbf{Average}} \\ \hline
BTF(Raw) &
  0.506 &
  0.523 &
  0.490 &
  0.464 &
  0.535 &
  {\color[HTML]{FE0000} 0.760} &
  0.633 &
  0.598 &
  {\color[HTML]{3531FF} 0.793} &
  {\color[HTML]{FE0000} 0.751} &
  {\color[HTML]{FE0000} 0.579} &
  0.700 &
  0.611 \\
BTF(FPFH) &
  0.659 &
  0.653 &
  0.638 &
  0.814 &
  0.677 &
  {\color[HTML]{3531FF} 0.620} &
  0.638 &
  0.603 &
  0.567 &
  0.434 &
  0.557 &
  0.505 &
  0.614 \\
M3DM(PointBERT) &
  0.497 &
  0.517 &
  0.480 &
  0.716 &
  0.661 &
  0.569 &
  0.628 &
  {\color[HTML]{3531FF} 0.628} &
  0.491 &
  0.638 &
  {\color[HTML]{3531FF} 0.573} &
  0.569 &
  0.581 \\
M3DM(PointMAE) &
  0.479 &
  0.508 &
  0.498 &
  0.739 &
  0.620 &
  0.533 &
  0.525 &
  {\color[HTML]{FE0000} 0.663} &
  0.518 &
  0.616 &
  {\color[HTML]{3531FF} 0.573} &
  0.593 &
  0.572 \\
PatchCore(FPFH) &
  {\color[HTML]{FE0000} 0.852} &
  0.611 &
  0.553 &
  0.872 &
  0.569 &
  0.506 &
  0.642 &
  0.411 &
  0.508 &
  0.573 &
  0.491 &
  0.506 &
  0.591 \\
PatchCore(FPFH+Raw) &
  {\color[HTML]{3531FF} 0.807} &
  {\color[HTML]{FE0000} 0.766} &
  0.611 &
  {\color[HTML]{FE0000} 0.885} &
  0.767 &
  0.560 &
  0.844 &
  0.411 &
  0.763 &
  0.553 &
  0.473 &
  0.559 &
  0.667 \\
PatchCore(PointMAE) &
  0.747 &
  0.555 &
  0.576 &
  0.864 &
  0.801 &
  0.488 &
  0.720 &
  0.444 &
  0.546 &
  0.590 &
  0.561 &
  0.708 &
  0.633 \\
\textbf{Reg3D-AD} &
  0.703 &
  {\color[HTML]{3531FF} 0.753} &
  {\color[HTML]{3531FF} 0.824} &
  {\color[HTML]{3531FF} 0.884} &
  {\color[HTML]{3531FF} 0.884} &
  0.588 &
  {\color[HTML]{3531FF} 0.939} &
  0.454 &
  0.787 &
  0.646 &
  0.491 &
  {\color[HTML]{3531FF} 0.721} &
  {\color[HTML]{3531FF} 0.723} \\
\textbf{Group3AD(Ours)} &
  0.757 &
  0.706 &
  {\color[HTML]{FE0000} 0.837} &
  0.674 &
  {\color[HTML]{FE0000} 0.932} &
  0.612 &
  {\color[HTML]{FE0000} 0.981} &
  0.533 &
  {\color[HTML]{FE0000} 0.842} &
  {\color[HTML]{3531FF} 0.648} &
  0.567 &
  {\color[HTML]{FE0000} 0.785} &
  {\color[HTML]{FE0000} 0.740} \\ \bottomrule
\end{tabular}}
\label{table2} 
\end{table*}

\begin{table*}[th]
\caption{P-AUROC score ($\uparrow$) on Real3D-AD. The best and second-best results are marked in red and blue, respectively.}
\resizebox{\textwidth}{!}{
\begin{tabular}{l|cccccccccccc|c}
\toprule
\textbf{Method} &
  \multicolumn{1}{l}{\textbf{Airplane}} &
  \multicolumn{1}{l}{\textbf{Car}} &
  \multicolumn{1}{l}{\textbf{Candybar}} &
  \multicolumn{1}{l}{\textbf{Chicken}} &
  \multicolumn{1}{l}{\textbf{Diamond}} &
  \multicolumn{1}{l}{\textbf{Duck}} &
  \multicolumn{1}{l}{\textbf{Fish}} &
  \multicolumn{1}{l}{\textbf{Gemstone}} &
  \multicolumn{1}{l}{\textbf{Seahorse}} &
  \multicolumn{1}{l}{\textbf{Shell}} &
  \multicolumn{1}{l}{\textbf{Starfish}} &
  \multicolumn{1}{l|}{\textbf{Toffees}} &
  \multicolumn{1}{l}{\textbf{Average}} \\ \hline
BTF(Raw) &
  0.564 &
  0.647 &
  0.735 &
  0.608 &
  0.563 &
  0.601 &
  0.514 &
  0.597 &
  0.520 &
  0.489 &
  0.392 &
  0.623 &
  0.571 \\
BTF(FPFH) &
  {\color[HTML]{FE0000} 0.738} &
  0.708 &
  {\color[HTML]{FE0000} 0.864} &
  0.693 &
  {\color[HTML]{FE0000} 0.882} &
  {\color[HTML]{FE0000} 0.875} &
  0.709 &
  {\color[HTML]{FE0000} 0.891} &
  0.512 &
  0.571 &
  0.501 &
  {\color[HTML]{3531FF} 0.815} &
  0.730 \\
M3DM(PointBERT) &
  0.523 &
  0.593 &
  0.682 &
  {\color[HTML]{FE0000} 0.790} &
  0.594 &
  0.668 &
  0.589 &
  0.646 &
  0.574 &
  0.732 &
  0.563 &
  0.677 &
  0.636 \\
M3DM(PointMAE) &
  0.530 &
  0.607 &
  0.683 &
  0.735 &
  0.618 &
  0.678 &
  0.600 &
  0.654 &
  0.561 &
  0.748 &
  0.555 &
  0.679 &
  0.637 \\
PatchCore(FPFH) &
  0.471 &
  0.643 &
  0.637 &
  0.618 &
  0.760 &
  0.430 &
  0.464 &
  {\color[HTML]{3531FF} 0.830} &
  0.544 &
  0.596 &
  0.522 &
  0.411 &
  0.577 \\
PatchCore(FPFH+Raw) &
  0.556 &
  0.740 &
  {\color[HTML]{3531FF} 0.749} &
  0.558 &
  0.854 &
  0.658 &
  0.781 &
  0.539 &
  0.808 &
  0.753 &
  0.613 &
  0.549 &
  0.680 \\
PatchCore(PointMAE) &
  0.579 &
  0.610 &
  0.635 &
  0.683 &
  0.776 &
  0.439 &
  0.714 &
  0.514 &
  0.660 &
  0.725 &
  {\color[HTML]{3531FF} 0.641} &
  0.727 &
  0.642 \\
CPMF &
  0.618 &
  {\color[HTML]{FE0000} 0.836} &
  0.734 &
  0.559 &
  0.753 &
  {\color[HTML]{3531FF} 0.719} &
  {\color[HTML]{FE0000} 0.988} &
  0.449 &
  {\color[HTML]{FE0000} 0.962} &
  0.725 &
  {\color[HTML]{FE0000} 0.800} &
  {\color[HTML]{FE0000} 0.959} &
  {\color[HTML]{FE0000} 0.758} \\
\textbf{Reg3D-AD} &
  0.631 &
  0.718 &
  0.724 &
  0.676 &
  0.835 &
  0.503 &
  0.826 &
  0.545 &
  0.817 &
  {\color[HTML]{FE0000} 0.811} &
  0.617 &
  0.759 &
  0.705 \\
\textbf{Group3AD(Ours)} &
  {\color[HTML]{3531FF} 0.636} &
  {\color[HTML]{3531FF} 0.745} &
  0.738 &
  {\color[HTML]{3531FF} 0.759} &
  {\color[HTML]{3531FF} 0.862} &
  0.631 &
  {\color[HTML]{3531FF} 0.836} &
  0.564 &
  {\color[HTML]{3531FF} 0.827} &
  {\color[HTML]{3531FF} 0.798} &
  0.625 &
  0.803 &
  {\color[HTML]{3531FF} 0.735} \\\bottomrule
\end{tabular}}
\label{table3} 
\end{table*}

\begin{table*}[th]
\caption{P-AUPR score ($\uparrow$) on Real3D-AD. The best and second-best results are marked in red and blue, respectively.}
\resizebox{\textwidth}{!}{
\begin{tabular}{l|cccccccccccc|c}
\toprule
\textbf{Method} &
  \multicolumn{1}{l}{\textbf{Airplane}} &
  \multicolumn{1}{l}{\textbf{Car}} &
  \multicolumn{1}{l}{\textbf{Candybar}} &
  \multicolumn{1}{l}{\textbf{Chicken}} &
  \multicolumn{1}{l}{\textbf{Diamond}} &
  \multicolumn{1}{l}{\textbf{Duck}} &
  \multicolumn{1}{l}{\textbf{Fish}} &
  \multicolumn{1}{l}{\textbf{Gemstone}} &
  \multicolumn{1}{l}{\textbf{Seahorse}} &
  \multicolumn{1}{l}{\textbf{Shell}} &
  \multicolumn{1}{l}{\textbf{Starfish}} &
  \multicolumn{1}{l|}{\textbf{Toffees}} &
  \multicolumn{1}{l}{\textbf{Average}} \\ \hline
BTF(Raw) &
  0.012 &
  0.014 &
  0.025 &
  0.049 &
  0.032 &
  0.020 &
  0.017 &
  0.014 &
  0.031 &
  0.011 &
  0.017 &
  0.016 &
  0.022 \\
BTF(FPFH) &
  {\color[HTML]{FE0000} 0.027} &
  0.028 &
  0.118 &
  0.044 &
  0.239 &
  {\color[HTML]{FE0000} 0.068} &
  0.036 &
  {\color[HTML]{3531FF} 0.075} &
  0.027 &
  0.018 &
  0.034 &
  0.055 &
  0.064 \\
M3DM(PointBERT) &
  0.007 &
  0.017 &
  0.016 &
  {\color[HTML]{FE0000} 0.377} &
  0.038 &
  0.011 &
  0.039 &
  0.017 &
  0.028 &
  0.021 &
  0.040 &
  0.018 &
  0.052 \\
M3DM(PointMAE) &
  0.007 &
  0.018 &
  0.016 &
  {\color[HTML]{3531FF} 0.310} &
  0.033 &
  0.011 &
  0.025 &
  0.018 &
  0.030 &
  0.022 &
  0.040 &
  0.021 &
  0.046 \\
PatchCore(FPFH) &
  {\color[HTML]{FE0000} 0.027} &
  0.034 &
  {\color[HTML]{FE0000} 0.142} &
  0.040 &
  0.273 &
  {\color[HTML]{3531FF} 0.055} &
  0.052 &
  {\color[HTML]{FE0000} 0.093} &
  0.031 &
  0.031 &
  0.037 &
  0.040 &
  0.071 \\
PatchCore(FPFH+Raw) &
  0.016 &
  {\color[HTML]{3531FF} 0.160} &
  0.092 &
  0.045 &
  {\color[HTML]{FE0000} 0.363} &
  0.034 &
  0.266 &
  0.066 &
  {\color[HTML]{3531FF} 0.291} &
  0.049 &
  0.035 &
  0.055 &
  0.123 \\
PatchCore(PointMAE) &
  0.016 &
  0.069 &
  0.020 &
  0.052 &
  0.107 &
  0.008 &
  0.201 &
  0.008 &
  0.071 &
  0.043 &
  0.046 &
  0.055 &
  0.058 \\
CPMF &
  0.010 &
  0.064 &
  0.050 &
  0.031 &
  0.074 &
  0.018 &
  {\color[HTML]{FE0000} 0.559} &
  0.007 &
  {\color[HTML]{FE0000} 0.636} &
  0.025 &
  {\color[HTML]{FE0000} 0.128} &
  {\color[HTML]{FE0000} 0.391} &
  {\color[HTML]{FE0000} 0.166} \\
\textbf{Reg3D-AD} &
  0.017 &
  0.135 &
  0.109 &
  0.044 &
  0.191 &
  0.010 &
  0.437 &
  0.016 &
  0.182 &
  {\color[HTML]{3531FF} 0.065} &
  0.039 &
  0.067 &
  0.109 \\
\textbf{Group3AD(Ours)} &
  {\color[HTML]{3531FF} 0.018} &
  {\color[HTML]{FE0000} 0.174} &
  {\color[HTML]{3531FF} 0.122} &
  0.068 &
  {\color[HTML]{3531FF} 0.287} &
  0.016 &
  {\color[HTML]{3531FF} 0.448} &
  0.009 &
  0.24 &
  {\color[HTML]{FE0000} 0.067} &
  {\color[HTML]{3531FF} 0.056} &
  {\color[HTML]{3531FF} 0.134} &
  {\color[HTML]{3531FF} 0.137} \\ \bottomrule
\end{tabular}}
\label{table4} 
\end{table*}
\begin{table*}[th]
\caption{Ablation studies on the INU $\&$ IAN Methods. The best and second-best results are marked in red and blue, respectively.}
\resizebox{\textwidth}{!}{
\begin{tabular}{c|lcccccccccccl|l}
\toprule
\multicolumn{1}{l|}{\textbf{Metric}} &
  \textbf{Method} &
  \multicolumn{1}{l}{\textbf{Airplane}} &
  \multicolumn{1}{l}{\textbf{Car}} &
  \multicolumn{1}{l}{\textbf{Candybar}} &
  \multicolumn{1}{l}{\textbf{Chicken}} &
  \multicolumn{1}{l}{\textbf{Diamond}} &
  \multicolumn{1}{l}{\textbf{Duck}} &
  \multicolumn{1}{l}{\textbf{Fish}} &
  \multicolumn{1}{l}{\textbf{Gemstone}} &
  \multicolumn{1}{l}{\textbf{Seahorse}} &
  \multicolumn{1}{l}{\textbf{Shell}} &
  \multicolumn{1}{l}{\textbf{Starfish}} &
  \multicolumn{1}{l}{\textbf{Toffees}} &
  \multicolumn{1}{|l}{\textbf{Average}} \\ \hline
\textbf{O-AUROC ($\uparrow$)} &
  PatchCore(PointMAE) &
  0.726 &
  0.498 &
  0.585 &
  {\color[HTML]{FE0000} 0.827} &
  0.783 &
  0.489 &
  0.630 &
  0.374 &
  0.539 &
  0.501 &
  0.519 &
  0.663 &
  0.594 \\
 &
  PatchCore(PointMAE) with IUN\&IAN &
  {\color[HTML]{FE0000} 0.748} &
  0.558 &
  0.656 &
  0.655 &
  0.845 &
  0.579 &
  0.741 &
  0.478 &
  0.556 &
  0.522 &
  0.53 &
  0.662 &
  0.628 \\
 &
  Group3AD without IUN\&IAN &
  0.721 &
  {\color[HTML]{3531FF} 0.712} &
  {\color[HTML]{3531FF} 0.842} &
  0.773 &
  {\color[HTML]{3531FF} 0.897} &
  {\color[HTML]{3531FF} 0.656} &
  {\color[HTML]{3531FF} 0.965} &
  {\color[HTML]{3531FF} 0.511} &
  {\color[HTML]{3531FF} 0.838} &
  {\color[HTML]{3531FF} 0.536} &
  {\color[HTML]{3531FF} 0.553} &
  {\color[HTML]{3531FF} 0.775} &
  {\color[HTML]{3531FF} 0.732} \\
 &
  \textbf{Group3AD} &
  {\color[HTML]{3531FF} 0.744} &
  {\color[HTML]{FE0000} 0.728} &
  {\color[HTML]{FE0000} 0.847} &
  {\color[HTML]{3531FF} 0.786} &
  {\color[HTML]{FE0000} 0.932} &
  {\color[HTML]{FE0000} 0.679} &
  {\color[HTML]{FE0000} 0.976} &
  {\color[HTML]{FE0000} 0.539} &
  {\color[HTML]{FE0000} 0.841} &
  {\color[HTML]{FE0000} 0.585} &
  {\color[HTML]{FE0000} 0.562} &
  {\color[HTML]{FE0000} 0.796} &
  {\color[HTML]{FE0000} 0.751} \\ \hline
\textbf{P-AUPR ($\uparrow$)} &
  PatchCore(PointMAE) &
  {\color[HTML]{3531FF} 0.016} &
  0.069 &
  0.020 &
  0.052 &
  0.107 &
  0.008 &
  0.201 &
  0.008 &
  0.071 &
  {\color[HTML]{3531FF} 0.043} &
  0.046 &
  0.055 &
  0.058 \\
 &
  PatchCore(PointMAE) with IUN\&IAN &
  0.013 &
  0.066 &
  0.021 &
  0.047 &
  0.118 &
  0.010 &
  0.210 &
  {\color[HTML]{3531FF} 0.009} &
  {\color[HTML]{3531FF} 0.081} &
  0.037 &
  {\color[HTML]{FE0000} 0.063} &
  0.12 &
  0.066 \\
 &
  Group3AD without IUN\&IAN &
  0.011 &
  {\color[HTML]{3531FF} 0.158} &
  {\color[HTML]{3531FF} 0.100} &
  {\color[HTML]{3531FF} 0.051} &
  {\color[HTML]{3531FF} 0.210} &
  {\color[HTML]{3531FF} 0.014} &
  {\color[HTML]{3531FF} 0.420} &
  {\color[HTML]{FE0000} 0.010} &
  {\color[HTML]{FE0000} 0.240} &
  0.037 &
  0.033 &
  0.101 &
  0.115 \\
 &
  \textbf{Group3AD} &
  {\color[HTML]{FE0000} 0.018} &
  {\color[HTML]{FE0000} 0.174} &
  {\color[HTML]{FE0000} 0.122} &
  {\color[HTML]{FE0000} 0.068} &
  {\color[HTML]{FE0000} 0.287} &
  {\color[HTML]{FE0000} 0.016} &
  {\color[HTML]{FE0000} 0.448} &
  {\color[HTML]{3531FF} 0.009} &
  {\color[HTML]{FE0000} 0.240} &
  {\color[HTML]{FE0000} 0.067} &
  {\color[HTML]{3531FF} 0.056} &
  {\color[HTML]{FE0000} 0.134} &
  {\color[HTML]{FE0000} 0.137} \\ \bottomrule
\end{tabular}}
\label{table5} 
\end{table*}

\begin{table*}[th]
\caption{Ablation studies on the AGCS scheme. The best and second-best results are marked in red and blue, respectively.}
\resizebox{\textwidth}{!}{
\begin{tabular}{c|lcccccccccccl|l}
\toprule
\multicolumn{1}{l|}{\textbf{Metric}} &
  \textbf{Method} &
  \multicolumn{1}{l}{\textbf{Airplane}} &
  \multicolumn{1}{l}{\textbf{Car}} &
  \multicolumn{1}{l}{\textbf{Candybar}} &
  \multicolumn{1}{l}{\textbf{Chicken}} &
  \multicolumn{1}{l}{\textbf{Diamond}} &
  \multicolumn{1}{l}{\textbf{Duck}} &
  \multicolumn{1}{l}{\textbf{Fish}} &
  \multicolumn{1}{l}{\textbf{Gemstone}} &
  \multicolumn{1}{l}{\textbf{Seahorse}} &
  \multicolumn{1}{l}{\textbf{Shell}} &
  \multicolumn{1}{l}{\textbf{Starfish}} &
  \multicolumn{1}{l}{\textbf{Toffees}} &
  \multicolumn{1}{|l}{\textbf{Average}} \\ \hline
\textbf{O-AUROC ($\uparrow$)} &
  PatchCore(PointMAE) &
  {\color[HTML]{3531FF} 0.726} &
  0.498 &
  0.585 &
  {\color[HTML]{FE0000} 0.827} &
  0.783 &
  0.489 &
  0.630 &
  0.374 &
  0.539 &
  0.501 &
  0.519 &
  0.663 &
  0.594 \\
 &
  PatchCore(PointMAE) with AGCS &
  0.713 &
  0.544 &
  0.614 &
  0.676 &
  0.820 &
  0.544 &
  0.752 &
  0.452 &
  0.590 &
  0.507 &
  0.524 &
  0.778 &
  0.626 \\
 &
  Group3AD without AGCS &
  0.723 &
  {\color[HTML]{3531FF} 0.708} &
  {\color[HTML]{3531FF} 0.833} &
  0.726 &
  {\color[HTML]{3531FF} 0.906} &
  {\color[HTML]{3531FF} 0.661} &
  {\color[HTML]{3531FF} 0.941} &
  {\color[HTML]{3531FF} 0.523} &
  {\color[HTML]{3531FF} 0.818} &
  {\color[HTML]{3531FF} 0.544} &
  {\color[HTML]{3531FF} 0.531} &
  {\color[HTML]{3531FF} 0.787} &
  {\color[HTML]{3531FF} 0.725} \\
 &
  \textbf{Group3AD} &
  {\color[HTML]{FE0000} 0.744} &
  {\color[HTML]{FE0000} 0.728} &
  {\color[HTML]{FE0000} 0.847} &
  {\color[HTML]{3531FF} 0.786} &
  {\color[HTML]{FE0000} 0.932} &
  {\color[HTML]{FE0000} 0.679} &
  {\color[HTML]{FE0000} 0.976} &
  {\color[HTML]{FE0000} 0.539} &
  {\color[HTML]{FE0000} 0.841} &
  {\color[HTML]{FE0000} 0.585} &
  {\color[HTML]{FE0000} 0.562} &
  {\color[HTML]{FE0000} 0.796} &
  {\color[HTML]{FE0000} 0.751} \\ \hline
\textbf{P-AUPR ($\uparrow$)} &
  PatchCore(PointMAE) &
  0.016 &
  0.069 &
  0.020 &
  0.052 &
  0.107 &
  0.008 &
  0.201 &
  {\color[HTML]{3531FF} 0.008} &
  0.071 &
  0.043 &
  0.046 &
  0.055 &
  0.058 \\
 &
  PatchCore(PointMAE) with AGCS &
  0.010 &
  0.057 &
  0.022 &
  {\color[HTML]{3531FF} 0.055} &
  0.145 &
  0.013 &
  0.261 &
  {\color[HTML]{FE0000} 0.009} &
  0.092 &
  0.040 &
  0.039 &
  {\color[HTML]{FE0000} 0.137} &
  0.073 \\
 &
  Group3AD without AGCS &
  {\color[HTML]{FE0000} 0.019} &
  {\color[HTML]{3531FF} 0.176} &
  {\color[HTML]{3531FF} 0.067} &
  0.058 &
  {\color[HTML]{3531FF} 0.253} &
  {\color[HTML]{3531FF} 0.015} &
  {\color[HTML]{3531FF} 0.437} &
  {\color[HTML]{3531FF} 0.008} &
  {\color[HTML]{3531FF} 0.237} &
  {\color[HTML]{3531FF} 0.048} &
  {\color[HTML]{3531FF} 0.042} &
  0.124 &
  {\color[HTML]{3531FF} 0.124} \\
 &
  \textbf{Group3AD} &
  {\color[HTML]{3531FF} 0.018} &
  {\color[HTML]{FE0000} 0.174} &
  {\color[HTML]{FE0000} 0.122} &
  {\color[HTML]{FE0000} 0.068} &
  {\color[HTML]{FE0000} 0.287} &
  {\color[HTML]{FE0000} 0.016} &
  {\color[HTML]{FE0000} 0.448} &
  {\color[HTML]{FE0000} 0.009} &
  {\color[HTML]{FE0000} 0.240} &
  {\color[HTML]{FE0000} 0.067} &
  {\color[HTML]{FE0000} 0.056} &
  {\color[HTML]{3531FF} 0.134} &
  {\color[HTML]{FE0000} 0.137} \\ \bottomrule
\end{tabular}}
\label{table6} 
\end{table*}

\section{EXPERIMENTS}
\subsection{Experimental Settings}

In our experiments, we utilize the Real3D-AD~\cite{liu2024real3d} dataset. The experimental setup and evaluation criteria are based on the Real3D-AD project. We assess the performance of anomaly detection at both the object and point levels using the Area Under the Receiver Operating Characteristic Curve (AUROC) and the Area Under the Precision-Recall Curve (AUPR).

\subsection{Results and Analysis}
\subsubsection{Anomaly Detection on Real3D-AD}
Tables~\ref{table1}-\ref{table4} summarize per-class comparisons between Group3AD and other state-of-the-art methods, namely 
Reg3D-AD~\cite{liu2024real3d}, CPMF~\cite{cao2023complementary}, IMRNet~\cite{Li_2024_CVPR}, and several benchmarking methods reported in Real3D-AD~\cite{liu2024real3d}.

(1) \textbf{O-AUROC}. Regarding the O-AUROC metric, while Reg3D-AD~\cite{liu2024real3d} sets the benchmark among current approaches with an average O-AUROC of 0.704, this mark falls short of being entirely effective. The introduced Group3AD method, however, markedly surpasses these existing standards, registering an impressive O-AUROC of 0.751, illustrating a significant advancement over prior techniques, as shown in Table~\ref{table1}.

(2) \textbf{O-AUPR}. In the context of O-AUPR, Reg3D-AD leads among contemporary strategies with an average performance of 0.723, indicating a gap towards optimal precision-recall balance. The newly developed Group3AD method exceeds these precedents, demonstrating a notable O-AUPR of 0.74, showcasing a substantial improvement in accurately identifying anomalies, as shown in Table~\ref{table2}.

(3) \textbf{P-AUROC}. Our findings indicate that Group3AD exhibits a marked improvement over Reg3D-AD in the context of P-AUROC scores across various test scenarios. While Reg3D-AD achieved a P-AUROC of 0.705, Group3AD advanced this benchmark to 0.735, indicating a significant enhancement in detection capability, as shown in Table~\ref{table3}. This improvement demonstrates the robustness of Group3AD in navigating the complexities of 3D anomaly detection.

(4) \textbf{P-AUPR}. Reflecting on the P-AUPR metric. Reg3D-AD sets a solid foundation with a P-AUPR score of 0.109. Our novel approach, Group3AD, significantly enhances this benchmark by achieving a P-AUPR of 0.137, indicating a marked improvement in recall capabilities, as shown in Table~\ref{table4}. 

In reviewing the performance across O-AUROC, O-AUPR, P-AUROC, and P-AUPR metrics, it is clear that while Reg3D-AD~\cite{liu2024real3d} has set solid benchmarks, Group3AD significantly surpasses these.

\subsubsection{Evaluating Intercluster Uniformity \& Intracluster Alignment Method for Anomaly Detection}

 By comparing the performance metrics of models with the IUN\&IAN component both enabled and disabled, From Table~\ref{table5}, we observed a significant enhancement in the model's ability to discriminate between normal and anomalous features when the IUN\&IAN was active. This enhancement directly correlates with IUN\&IAN's primary function: to optimize the separation and uniform distribution of feature clusters in the feature space. While the IUN\&IAN establishes a foundational separation and distribution of feature clusters, the IAN gently fine-tunes this landscape, promoting tighter and more cohesive clusters. Such optimization evidently aids in reducing ambiguity and overlap between clusters, thereby sharpening the model's anomaly detection capabilities.

\subsubsection{Evaluating Adaptive Group-Center Selection for Anomaly Detection}
We further examined the role of the AGCS within Group3AD. Experimental results shown in Table~\ref{table6} indicate that AGCS substantially improves the model's effectiveness in detecting anomalies. By concentrating on areas with potential irregularities, AGCS allows the model to identify subtle yet critical anomalies that might be missed under uniform sampling conditions. 

\begin{figure}[htbp]
    \centering
    \includegraphics[width=0.75\linewidth]{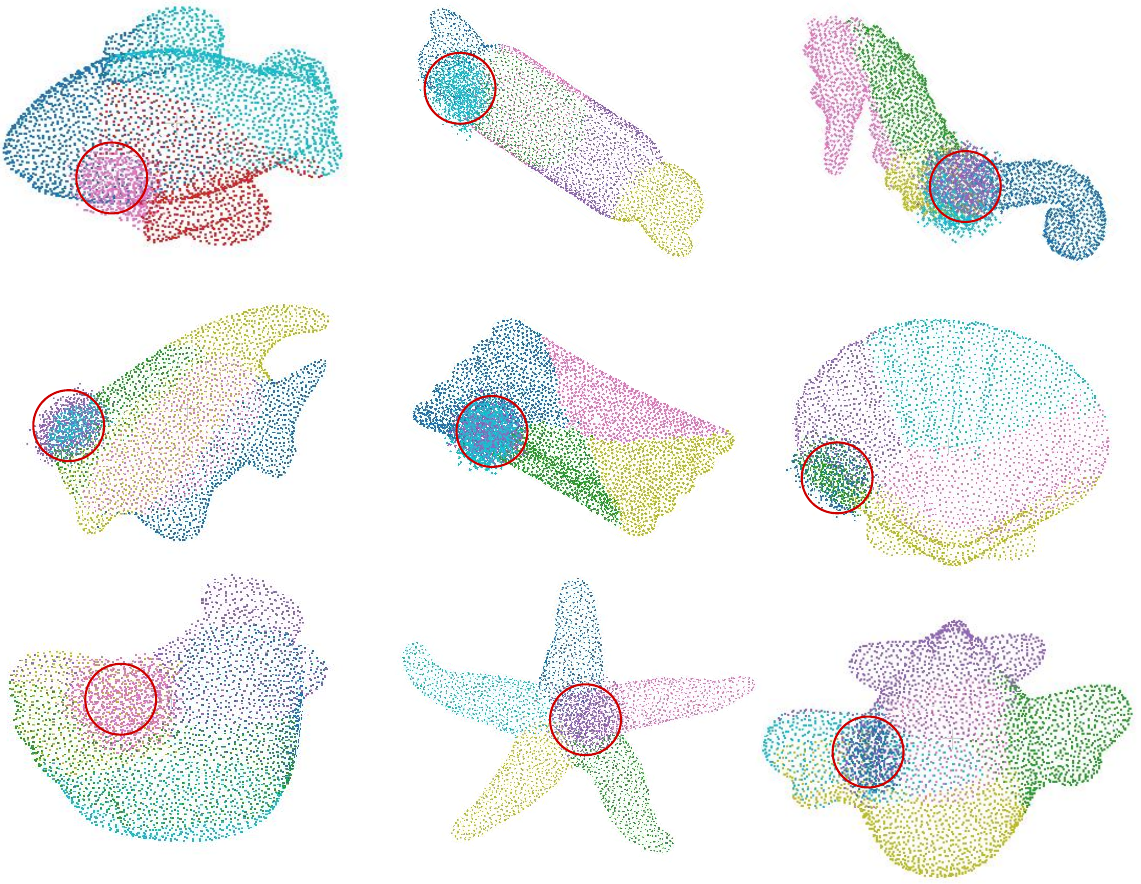}
\caption{Visualization results obtained by Group3AD. Different colors indicate different groups selected by AGCS. The red circle represents anomaly area.}
    \label{fig:vis_defect}
\end{figure}

\subsection{Visualization}
Figure~\ref{fig:vis_defect} provides a visual representation of the clustering results achieved with the Group3AD framework on the Real3D-AD dataset. Each dot signifies a group center, pinpointed through the refined feature space sculpted by the IUN and IAN. The clear separation of clusters showcases the IUN's impact in enhancing feature discrimination across groups, while the density of points within each cluster attests to the IAN's role in fostering coherence among features.


\section{Conclusions}
In this work, we introduced Group3AD, a robust framework tailored for high-resolution 3D-AD in industry, which can effectively enhance anomaly detection and localization accuracy. Demonstrating significant improvements over existing methodologies, especially in terms of reducing false positives and leveraging depth information for clearer anomaly identification, Group3AD emerges as a practical solution for real-world applications. Future endeavors may extend its applicability and efficiency, further solidifying its utility in industrial anomaly detection tasks.



\section{Acknowledgments}
This work was supported by the National Natural Science Foundation of China (Grant Nos. 62206122, 82261138629, 62302309), the Guangdong Provincial Key Laboratory (Grant No. 2023B1212060076), Shenzhen Municipal Science and Technology Innovation Council (Grant No. JCYJ20220531101412030), Tencent ``Rhinoceros Birds” - Scientific Research Foundation for Young Teachers of Shenzhen University, and the Internal Fund of National Engineering Laboratory for Big Data System Computing Technology (Grant No. SZU-BDSC-IF2024-08).
\bibliographystyle{ACM-Reference-Format}
\bibliography{sample-base}

\begin{thebibliography}{10}
\providecommand{\url}[1]{#1}
\csname url@samestyle\endcsname
\providecommand{\newblock}{\relax}
\providecommand{\bibinfo}[2]{#2}
\providecommand{\BIBentrySTDinterwordspacing}{\spaceskip=0pt\relax}
\providecommand{\BIBentryALTinterwordstretchfactor}{4}
\providecommand{\BIBentryALTinterwordspacing}{\spaceskip=\fontdimen2\font plus
\BIBentryALTinterwordstretchfactor\fontdimen3\font minus \fontdimen4\font\relax}
\providecommand{\BIBforeignlanguage}[2]{{%
\expandafter\ifx\csname l@#1\endcsname\relax
\typeout{** WARNING: IEEEtran.bst: No hyphenation pattern has been}%
\typeout{** loaded for the language `#1'. Using the pattern for}%
\typeout{** the default language instead.}%
\else
\language=\csname l@#1\endcsname
\fi
#2}}
\providecommand{\BIBdecl}{\relax}
\BIBdecl

\bibitem{liu2024real3d}
J.~Liu, G.~Xie, R.~Chen, X.~Li, J.~Wang, Y.~Liu, C.~Wang, and F.~Zheng, ``Real3d-ad: A dataset of point cloud anomaly detection,'' \emph{Advances in Neural Information Processing Systems}, vol.~36, 2024.

\bibitem{liu2023real3d}
J.~Liu, G.~Xie, X.~Li, J.~Wang, Y.~Liu, C.~Wang, F.~Zheng \emph{et~al.}, ``Real3d-ad: A dataset of point cloud anomaly detection,'' in \emph{Thirty-seventh Conference on Neural Information Processing Systems Datasets and Benchmarks Track}, 2023.

\bibitem{xie2024iad}
G.~Xie, J.~Wang, J.~Liu, J.~Lyu, Y.~Liu, C.~Wang, F.~Zheng, and Y.~Jin, ``Im-iad: Industrial image anomaly detection benchmark in manufacturing,'' \emph{IEEE Transactions on Cybernetics}, 2024.

\bibitem{li2022towards}
W.~Li, J.~Zhan, J.~Wang, B.~Xia, B.-B. Gao, J.~Liu, C.~Wang, and F.~Zheng, ``Towards continual adaptation in industrial anomaly detection,'' in \emph{Proceedings of the 30th ACM International Conference on Multimedia}, 2022, pp. 2871--2880.

\bibitem{liu2024deep}
J.~Liu, G.~Xie, J.~Wang, S.~Li, C.~Wang, F.~Zheng, and Y.~Jin, ``Deep industrial image anomaly detection: A survey,'' \emph{Machine Intelligence Research}, vol.~21, no.~1, pp. 104--135, 2024.

\bibitem{xie2023pushing}
G.~Xie, J.~Wang, J.~Liu, F.~Zheng, and Y.~Jin, ``Pushing the limits of fewshot anomaly detection in industry vision: Graphcore,'' \emph{arXiv preprint arXiv:2301.12082}, 2023.

\bibitem{cao2023complementary}
Y.~Cao, X.~Xu, and W.~Shen, ``Complementary pseudo multimodal feature for point cloud anomaly detection,'' \emph{arXiv preprint arXiv:2303.13194}, 2023.

\bibitem{chen2023easynet}
R.~Chen, G.~Xie, J.~Liu, J.~Wang, Z.~Luo, J.~Wang, and F.~Zheng, ``Easynet: An easy network for 3d industrial anomaly detection,'' in \emph{Proceedings of the 31st ACM International Conference on Multimedia}, 2023, pp. 7038--7046.

\bibitem{gao2019representation}
J.~Gao, D.~He, X.~Tan, T.~Qin, L.~Wang, and T.-Y. Liu, ``Representation degeneration problem in training natural language generation models,'' \emph{arXiv preprint arXiv:1907.12009}, 2019.

\bibitem{li2023towards}
W.~Li and X.~Xu, ``Towards scalable 3d anomaly detection and localization: A benchmark via 3d anomaly synthesis and a self-supervised learning network,'' \emph{arXiv preprint arXiv:2311.14897}, 2023.

\bibitem{bergmann2020uninformed}
P.~Bergmann, M.~Fauser, D.~Sattlegger, and C.~Steger, ``Uninformed students: Student-teacher anomaly detection with discriminative latent embeddings,'' in \emph{Proceedings of the IEEE/CVF conference on computer vision and pattern recognition}, 2020, pp. 4183--4192.

\bibitem{bergmann2021mvtec}
P.~Bergmann, X.~Jin, D.~Sattlegger, and C.~Steger, ``The mvtec 3d-ad dataset for unsupervised 3d anomaly detection and localization,'' \emph{arXiv preprint arXiv:2112.09045}, 2021.

\bibitem{jiang2022softpatch}
X.~Jiang, J.~Liu, J.~Wang, Q.~Nie, K.~Wu, Y.~Liu, C.~Wang, and F.~Zheng, ``Softpatch: Unsupervised anomaly detection with noisy data,'' \emph{Advances in Neural Information Processing Systems}, vol.~35, pp. 15\,433--15\,445, 2022.

\bibitem{defard2021padim}
T.~Defard, A.~Setkov, A.~Loesch, and R.~Audigier, ``Padim: a patch distribution modeling framework for anomaly detection and localization,'' in \emph{International Conference on Pattern Recognition}.\hskip 1em plus 0.5em minus 0.4em\relax Springer, 2021, pp. 475--489.

\bibitem{wang2019improving}
L.~Wang, J.~Huang, K.~Huang, Z.~Hu, G.~Wang, and Q.~Gu, ``Improving neural language generation with spectrum control,'' in \emph{International Conference on Learning Representations}, 2019.

\bibitem{ethayarajh2019contextual}
K.~Ethayarajh, ``How contextual are contextualized word representations? comparing the geometry of bert, elmo, and gpt-2 embeddings,'' \emph{arXiv preprint arXiv:1909.00512}, 2019.

\bibitem{afham2022crosspoint}
M.~Afham, I.~Dissanayake, D.~Dissanayake, A.~Dharmasiri, K.~Thilakarathna, and R.~Rodrigo, ``Crosspoint: Self-supervised cross-modal contrastive learning for 3d point cloud understanding,'' in \emph{Proceedings of the IEEE/CVF Conference on Computer Vision and Pattern Recognition}, 2022, pp. 9902--9912.

\bibitem{liu2023fac}
K.~Liu, A.~Xiao, X.~Zhang, S.~Lu, and L.~Shao, ``Fac: 3d representation learning via foreground aware feature contrast,'' in \emph{Proceedings of the IEEE/CVF Conference on Computer Vision and Pattern Recognition}, 2023, pp. 9476--9485.

\bibitem{gao2021simcse}
T.~Gao, X.~Yao, and D.~Chen, ``Simcse: Simple contrastive learning of sentence embeddings,'' \emph{arXiv preprint arXiv:2104.08821}, 2021.

\bibitem{cai2020isotropy}
X.~Cai, J.~Huang, Y.~Bian, and K.~Church, ``Isotropy in the contextual embedding space: Clusters and manifolds,'' in \emph{International conference on learning representations}, 2020.

\bibitem{bergmann2019mvtec}
P.~Bergmann, M.~Fauser, D.~Sattlegger, and C.~Steger, ``Mvtec ad--a comprehensive real-world dataset for unsupervised anomaly detection,'' in \emph{Proceedings of the IEEE/CVF conference on computer vision and pattern recognition}, 2019, pp. 9592--9600.

\bibitem{roth2022towards}
K.~Roth, L.~Pemula, J.~Zepeda, B.~Sch{\"o}lkopf, T.~Brox, and P.~Gehler, ``Towards total recall in industrial anomaly detection,'' in \emph{Proceedings of the IEEE/CVF Conference on Computer Vision and Pattern Recognition}, 2022, pp. 14\,318--14\,328.

\bibitem{li2021cutpaste}
C.-L. Li, K.~Sohn, J.~Yoon, and T.~Pfister, ``Cutpaste: Self-supervised learning for anomaly detection and localization,'' in \emph{Proceedings of the IEEE/CVF conference on computer vision and pattern recognition}, 2021, pp. 9664--9674.

\bibitem{liu2023simplenet}
Z.~Liu, Y.~Zhou, Y.~Xu, and Z.~Wang, ``Simplenet: A simple network for image anomaly detection and localization,'' in \emph{Proceedings of the IEEE/CVF Conference on Computer Vision and Pattern Recognition}, 2023, pp. 20\,402--20\,411.

\bibitem{deng2022anomaly}
H.~Deng and X.~Li, ``Anomaly detection via reverse distillation from one-class embedding,'' in \emph{Proceedings of the IEEE/CVF Conference on Computer Vision and Pattern Recognition}, 2022, pp. 9737--9746.

\bibitem{gudovskiy2022cflow}
D.~Gudovskiy, S.~Ishizaka, and K.~Kozuka, ``Cflow-ad: Real-time unsupervised anomaly detection with localization via conditional normalizing flows,'' in \emph{Proceedings of the IEEE/CVF winter conference on applications of computer vision}, 2022, pp. 98--107.

\bibitem{rudolph2021same}
M.~Rudolph, B.~Wandt, and B.~Rosenhahn, ``Same same but differnet: Semi-supervised defect detection with normalizing flows,'' in \emph{Proceedings of the IEEE/CVF winter conference on applications of computer vision}, 2021, pp. 1907--1916.

\bibitem{zavrtanik2021draem}
V.~Zavrtanik, M.~Kristan, and D.~Sko{\v{c}}aj, ``Draem-a discriminatively trained reconstruction embedding for surface anomaly detection,'' in \emph{Proceedings of the IEEE/CVF International Conference on Computer Vision}, 2021, pp. 8330--8339.

\bibitem{wang2021student}
G.~Wang, S.~Han, E.~Ding, and D.~Huang, ``Student-teacher feature pyramid matching for anomaly detection,'' \emph{arXiv preprint arXiv:2103.04257}, 2021.

\bibitem{schluter2022natural}
H.~M. Schl{\"u}ter, J.~Tan, B.~Hou, and B.~Kainz, ``Natural synthetic anomalies for self-supervised anomaly detection and localization,'' in \emph{European Conference on Computer Vision}.\hskip 1em plus 0.5em minus 0.4em\relax Springer, 2022, pp. 474--489.

\bibitem{zavrtanik2022dsr}
V.~Zavrtanik, M.~Kristan, and D.~Sko{\v{c}}aj, ``Dsr--a dual subspace re-projection network for surface anomaly detection,'' in \emph{European conference on computer vision}.\hskip 1em plus 0.5em minus 0.4em\relax Springer, 2022, pp. 539--554.

\bibitem{salehi2021multiresolution}
M.~Salehi, N.~Sadjadi, S.~Baselizadeh, M.~H. Rohban, and H.~R. Rabiee, ``Multiresolution knowledge distillation for anomaly detection,'' in \emph{Proceedings of the IEEE/CVF conference on computer vision and pattern recognition}, 2021, pp. 14\,902--14\,912.

\bibitem{rudolph2023asymmetric}
M.~Rudolph, T.~Wehrbein, B.~Rosenhahn, and B.~Wandt, ``Asymmetric student-teacher networks for industrial anomaly detection,'' in \emph{Proceedings of the IEEE/CVF winter conference on applications of computer vision}, 2023, pp. 2592--2602.

\bibitem{bergmann2023anomaly}
P.~Bergmann and D.~Sattlegger, ``Anomaly detection in 3d point clouds using deep geometric descriptors,'' in \emph{Proceedings of the IEEE/CVF Winter Conference on Applications of Computer Vision}, 2023, pp. 2613--2623.

\bibitem{horwitz2023back}
E.~Horwitz and Y.~Hoshen, ``Back to the feature: classical 3d features are (almost) all you need for 3d anomaly detection,'' in \emph{Proceedings of the IEEE/CVF Conference on Computer Vision and Pattern Recognition}, 2023, pp. 2967--2976.

\bibitem{wang2023multimodal}
Y.~Wang, J.~Peng, J.~Zhang, R.~Yi, Y.~Wang, and C.~Wang, ``Multimodal industrial anomaly detection via hybrid fusion,'' in \emph{Proceedings of the IEEE/CVF Conference on Computer Vision and Pattern Recognition}, 2023, pp. 8032--8041.

\bibitem{bonfiglioli2022eyecandies}
L.~Bonfiglioli, M.~Toschi, D.~Silvestri, N.~Fioraio, and D.~De~Gregorio, ``The eyecandies dataset for unsupervised multimodal anomaly detection and localization,'' in \emph{Proceedings of the Asian Conference on Computer Vision}, 2022, pp. 3586--3602.

\bibitem{Zhu_2024}
\BIBentryALTinterwordspacing
H.~Zhu and T.~Fang, ``Compound fault diagnosis of rolling bearings with few-shot based on dcgan-replknet,'' \emph{Measurement Science and Technology}, vol.~35, no.~6, p. 066105, mar 2024. [Online]. Available: \url{https://dx.doi.org/10.1088/1361-6501/ad24b5}
\BIBentrySTDinterwordspacing

\bibitem{li2022simipu}
Z.~Li, Z.~Chen, A.~Li, L.~Fang, Q.~Jiang, X.~Liu, J.~Jiang, B.~Zhou, and H.~Zhao, ``Simipu: Simple 2d image and 3d point cloud unsupervised pre-training for spatial-aware visual representations,'' in \emph{Proceedings of the AAAI Conference on Artificial Intelligence}, vol.~36, no.~2, 2022, pp. 1500--1508.

\bibitem{rusu2009fast}
R.~B. Rusu, N.~Blodow, and M.~Beetz, ``Fast point feature histograms (fpfh) for 3d registration,'' in \emph{2009 IEEE international conference on robotics and automation}.\hskip 1em plus 0.5em minus 0.4em\relax IEEE, 2009, pp. 3212--3217.

\bibitem{Li_2024_CVPR}
W.~Li, X.~Xu, Y.~Gu, B.~Zheng, S.~Gao, and Y.~Wu, ``Towards scalable 3d anomaly detection and localization: A benchmark via 3d anomaly synthesis and a self-supervised learning network,'' in \emph{Proceedings of the IEEE/CVF Conference on Computer Vision and Pattern Recognition (CVPR)}, June 2024, pp. 22\,207--22\,216.

\bibitem{macqueen1967some}
J.~MacQueen \emph{et~al.}, ``Some methods for classification and analysis of multivariate observations,'' in \emph{Proceedings of the fifth Berkeley symposium on mathematical statistics and probability}, vol.~1, no.~14.\hskip 1em plus 0.5em minus 0.4em\relax Oakland, CA, USA, 1967, pp. 281--297.

\bibitem{syakur2018integration}
M.~Syakur, B.~K. Khotimah, E.~Rochman, and B.~D. Satoto, ``Integration k-means clustering method and elbow method for identification of the best customer profile cluster,'' in \emph{IOP conference series: materials science and engineering}, vol. 336.\hskip 1em plus 0.5em minus 0.4em\relax IOP Publishing, 2018, p. 012017.

\bibitem{guo2020deep}
Y.~Guo, H.~Wang, Q.~Hu, H.~Liu, L.~Liu, and M.~Bennamoun, ``Deep learning for 3d point clouds: A survey,'' \emph{IEEE transactions on pattern analysis and machine intelligence}, vol.~43, no.~12, pp. 4338--4364, 2020.

\bibitem{lei2023pyramidflow}
J.~Lei, X.~Hu, Y.~Wang, and D.~Liu, ``Pyramidflow: High-resolution defect contrastive localization using pyramid normalizing flow,'' in \emph{Proceedings of the IEEE/CVF Conference on Computer Vision and Pattern Recognition}, 2023, pp. 14\,143--14\,152.

\end{thebibliography}

\appendix
\clearpage 
\begin{center}
\textbf{SUPPLEMENTARY}
\end{center}
\section{Real3D-AD Dataset}
Real3D-AD, comprising 1,254 high-resolution 3D items with each ranging from forty thousand to millions of points, stands as the most extensive dataset for high-precision 3D industrial anomaly detection currently available. Real3D-AD exceeds other available datasets for 3D anomaly detection in terms of point cloud resolution (0.0010mm-0.0015mm), comprehensive 360-degree coverage, and flawless prototype quality. To ensure both reproducibility and accessibility, the Real3D-AD dataset, benchmark source code, and Reg3D-AD are made available on the website at: ~\href{https://github.com/M-3LAB/Real3D-AD}{https://github.com/M-3LAB/Real3D-AD}.

The Real3D-AD dataset includes statistical data across 12 diverse categories. Each category's training set contains only four samples, reflecting a few-shot learning approach akin to 2D anomaly detection. Categories span a variety of toy manufacturing lines such as Airplane, Candybar, and Diamond, among others. The training samples offer comprehensive models of 3D objects, whereas the test samples are only scanned from one side. The dataset's characteristics, including a low ratio of anomaly points and a focus on transparency, make it particularly challenging yet suitable for point cloud anomaly detection tasks.

\section{Evaluation Metrics for Group3AD}
In anomaly detection, particularly in the specialized domain of high-resolution 3D anomaly detection, the precision in evaluating a model’s effectiveness is crucial for determining its utility in real-world applications. The Group3AD framework, specifically engineered for this purpose, incorporates a suite of advanced metrics tailored to accurately gauge the model's proficiency in detecting and pinpointing anomalies within detailed 3D environments. This section provides a foundational overview of the evaluation metrics used in Group3AD, detailing their theoretical basis and the significance they hold in enhancing the model's diagnostic capabilities.

These metrics are integral to verifying the robustness and reliability of Group3AD in various industrial and technological contexts, where precise anomaly localization can significantly impact operational safety and efficiency. The selection of these metrics is grounded in their ability to provide a comprehensive assessment of performance across different aspects of anomaly detection, from general detection accuracy to specific localization precision, ensuring that Group3AD meets the rigorous demands of high-stakes applications.

\subsection{O-AUROC (Object-level Area Under the Receiver Operating Characteristic Curve)}
The Receiver Operating Characteristic (ROC) curve is a graphical plot that illustrates the diagnostic ability of a binary classifier system as its discrimination threshold is varied. The curve is created by plotting the true positive rate (TPR, sensitivity) against the false positive rate (FPR, 1 - specificity) at various threshold settings. The AUROC represents the area under the ROC curve and provides an aggregate measure of performance across all possible classification thresholds. A model with perfect discrimination has an AUROC of 1.0, indicating it can perfectly differentiate between the classes across all thresholds.

O-AUROC assesses the overall ability of the model to distinguish between normal and anomalous objects as complete entities. It is particularly effective for evaluating performance in datasets with varied anomaly presence, ensuring the model's robustness in practical scenarios.

\subsection{P-AUROC (Point-level Area Under the Receiver Operating Characteristic Curve)}
Similar to O-AUROC, the P-AUROC focuses on the model’s performance at a granular level—each individual point in a 3D point cloud. This metric is crucial for applications requiring detailed analysis within objects, where each point's classification directly impacts the overall utility of the detection system.

P-AUROC provides a measure of the model's precision in classifying each point within an object, essential for tasks requiring fine-grained anomaly localization, such as in manufacturing or structural integrity assessments where small anomalies could signify significant defects.

\subsection{O-AUPR (Object-level Area Under the Precision-Recall Curve)}
The Precision-Recall (PR) curve shows the trade-off between precision and recall for different thresholds. Unlike the ROC curve, a PR curve provides a more informative picture of an algorithm's performance when the classes are very imbalanced. The area under the PR curve (AUPR) is particularly useful as it gives a single measure of performance that considers both the precision and the recall of the predictive model, making it ideal for evaluating models in skewed datasets.

O-AUPR is crucial for assessing Group3AD’s efficacy in contexts where anomalies are rare or very subtle, ensuring the model does not overwhelmingly misclassify normal objects as anomalies—a common issue in highly imbalanced datasets.

\subsection{P-AUPR (Point-level Area Under the Precision-Recall Curve)}
P-AUPR extends the concept of O-AUPR to the granularity of point-level evaluations within each object. This metric is particularly important when precise anomaly localization is crucial, and where false positives (normal points misclassified as anomalies) could lead to unnecessary actions, such as further inspections or repairs.

Together, these metrics provide a robust framework for evaluating the Group3AD system, ensuring it meets both the sensitivity and specificity requirements essential for practical deployments in industrial settings. These metrics not only assess the effectiveness of anomaly detection and localization but also help in tuning the model to improve its performance across diverse operational scenarios.

\begin{table*}[h!]
\setlength\tabcolsep{6pt}
\caption{O-AUROC score on MVTec 3D PCD.}
\begin{tabular*}{\linewidth}{c|ccccccccccc}
\toprule
\textbf{Method} & \textbf{Bagel} & \textbf{Cable Gland} & \textbf{Carrot} & \textbf{Cookie} & \textbf{Dowel} & \textbf{Foam} & \textbf{Peach} & \textbf{Potato} & \textbf{Rope} & \textbf{Tire} & \textbf{Mean} \\\hline
M3DM(PointMAE) & 0.985 & 0.673 & 0.958 & 0.949 & 0.843 & 0.756 & 0.860 & 0.925 & 0.899 & 0.834 & 0.868 \\
\textbf{Ours} & \textbf{0.989} & \textbf{0.722} & \textbf{0.976} & \textbf{0.954} &\textbf{ 0.852} & \textbf{0.769} & \textbf{0.910} & \textbf{0.940} & \textbf{0.909} & \textbf{0.864} & \textbf{0.889} \\ \bottomrule
\end{tabular*}
\label{b}
\end{table*}

\begin{table*}[h!]
\setlength\tabcolsep{4pt}
\caption{O-AUROC score on Anomaly-ShapeNet.}
\begin{tabular*}{\linewidth}{c|ccccccccccccc}
\toprule
\textbf{Method} & \textbf{cabinet0} & \textbf{cap1} & \textbf{cap2} & \textbf{chair0} & \textbf{cup2} & \textbf{desk0} & \textbf{knife0} & \textbf{knife1} & \textbf{microphone1} & \textbf{screen0} & \textbf{vase10} & \textbf{vase6} & \textbf{Mean} \\\hline
Reg3D-AD & 0.533 & \textbf{0.554} & 0.578 & 0.489 & 0.733 & 0.482 & 0.433 & 0.526 & 0.637 & 0.396 & 0.530 & 0.512 & 0.534 \\
\textbf{Ours} & \textbf{0.630} & 0.536 & \textbf{0.628} & \textbf{0.59} & \textbf{0.762} & \textbf{0.49} & \textbf{0.46} & \textbf{0.573} & \textbf{0.667} & \textbf{0.429} & \textbf{0.603} & \textbf{0.552} & \textbf{0.577} \\ \bottomrule
\end{tabular*}
\label{c}
\end{table*}

\section{Implementation Details}
Building upon the foundation established by the benchmark method Reg3D-AD, the Group3AD framework introduces significant enhancements to the encoder training process to better adapt to high-resolution 3D anomaly detection tasks. Similar to Reg3D-AD, the memory\_size of Group3AD was set to 10000 in the experiment. The other parameters and abnormal score calculation method of Group3AD in the experiment are also consistent with Reg3D-AD to ensure fairness.

IUN\&IAN used to reinforce the encoder pre trained with PointMAE. Taking airplanes as an example. Each airplane point cloud is divided into 4096 groups during the training process, each containing 128 neighboring points. Our method has 524288 available points, far exceeding the 8192 points used in conventional PointMAE pre training. Our method more effectively utilizes the information contained in high-resolution point cloud data, avoiding the waste of anomaly point cloud information in the downsampling process.
\begin{table}[!ht]
    \centering
    \caption{Ablation experiment on {$\alpha$} size}
    \begin{tabular}{cc}
    \toprule
        \textbf{$\alpha$} & \textbf{Enhanced AUROC} \\ \hline 
        0.1 & 0.022 \\ 
        0.2 & 0.028 \\ 
        0.3 & 0.014 \\     \toprule
        \label{a}
    \end{tabular}

\end{table}
AGCS prioritizes areas likely to contain anomalies by analyzing local geometric variations within the point clouds. In experiments, the center points obtained by AGCS occupied 20\% of the total center points, indicating that the system focused more intensively on these areas, improving the sensitivity and accuracy of the anomaly detection process. In the training process of IUN\&IAN, the original point cloud selects the center point through FPS, without using our proposed AGCS module to obtain a more uniform feature distribution in the memorybank. The AGCS is implemented as detailed in Algorithm~\ref{alg:agcs}. The ablation experiments in the Table~\ref{a} demonstrate that setting alpha to 0.2 yields the best results under the current experimental conditions.

\begin{algorithm}[ht]
\caption{Adaptive Group-Center Selection~(AGCS)}\label{alg:agcs}
\begin{algorithmic}[1]
\State \textbf{Input:} Point clouds $\mathbf{P}$, number of points $N$, attention factor $\alpha$
\State \textbf{Output:} Set of center points $\mathbf{C}_{\text{points}}$
\Function{AGCS}{$\mathbf{P}$, $N$, $\alpha$}
    \State $\mathbf{N} \gets$ ComputeNormals($\mathbf{P}$) \Comment{Compute surface normals for  each point in the point clouds $\mathbf{P}$}
    \State $\mathbf{F}_{\text{FPFH}} \gets$ ComputeFPFH($\mathbf{P}$, $\mathbf{N}$) \Comment{Compute FPFH features for the point clouds based on the surface normals}
    \State $\mathbf{F}_{\text{var}} \gets$ ComputeLocalVariation($\mathbf{F}_{\text{FPFH}}$) \Comment{Evaluate local geometric variation, indicative of potential anomalies}
    \State $\mathbf{P}_{\text{high\_var}} \gets$ SelectHighVariationPoints($\mathbf{F}_{\text{var}}$, $\alpha$)
    \Comment{Select points with high variation in their local geometric features, which are more likely to be near anomalies}
    \State $\mathbf{C}_{\text{FPFH}} \gets$ FPS($\mathbf{P}_{\text{high\_var}}$, $N \times \alpha$) \Comment{Sample using FPS on high variation points}
    \State $\mathbf{C}_{\text{FPS}} \gets$ FPS($\mathbf{P}$, $N \times (1 - \alpha)$) \Comment{Sample remaining points uniformly}
    \State $\mathbf{C}_{\text{points}} \gets \mathbf{C}_{\text{FPFH}} \cup \mathbf{C}_{\text{FPS}}$
    \State \textbf{return} $\mathbf{C}_{\text{points}}$
\EndFunction
\end{algorithmic}
\end{algorithm}

\section{Supplementary testing on other datasets}

Real3D-AD~\cite{liu2024real3d} was released on NeurIPS 2023 and is currently the only high-resolution point cloud dataset available. Adding more experiments can better demonstrate the effectiveness of Group3AD. Although they are not HRPCD datasets. We supplemented the experiment on \href{https://www.mvtec.com/company/research/datasets/mvtec-3d-ad}{MVTec 3D PCD} and \href{https://github.com/Chopper-233/Anomaly-ShapeNet}{Anomaly-ShapeNet-new}. Experiments on Anomaly ShapeNet~\cite{Li_2024_CVPR} were conducted using 12 newly released samples. MVTec 3D~\cite{bergmann2021mvtec} is an RGBD (2.5D) dataset, so Group3AD cannot be directly applied. Therefore, we apply IUN and IAN to the CVPR2023 SOTA method \href{https://github.com/nomewang/M3DM}{M3DM} to demonstrate their effectiveness on  MVTec 3D PCD. The results in Tables~\ref{b} and Tables~\ref{c} demonstrate the effectiveness of Group3AD.

\end{document}